%% file: acl_latex.tex
\pdfoutput=1

\documentclass[11pt]{article}

\usepackage[final]{acl}

\usepackage{times}
\usepackage{latexsym}
\usepackage{colortbl}
\usepackage{enumitem}

\usepackage{multirow} 
\usepackage{xtab}
\usepackage{graphicx}
\usepackage{longtable}
\usepackage{colortbl} 
\usepackage{booktabs} 

\usepackage{xcolor}      
\usepackage{tcolorbox} 
\newtcolorbox{questionbox}[1][]{
  colframe=black,        
  colback=white,         
  boxrule=0.5mm,         
  sharp corners,         
  fonttitle=\bfseries,   
  title=Question,        
  #1
}
\newtcolorbox{answerbox-pakistan}[1][]{
  colframe=black,        
  colback=white,         
  boxrule=0.5mm,         
  sharp corners,         
  fonttitle=\bfseries,   
  title=Answer: Pakistan,        
  #1
}

\newtcolorbox{answerbox-japan}[1][]{
  colframe=black,        
  colback=white,         
  boxrule=0.5mm,         
  sharp corners,         
  fonttitle=\bfseries,   
  title=Answer: Japan,        
  #1
}

\newtcolorbox{answerbox-britain}[1][]{
  colframe=black,        
  colback=white,         
  boxrule=0.5mm,         
  sharp corners,         
  fonttitle=\bfseries,   
  title=Answer: Britain,        
  #1
}

\definecolor{lightgray}{gray}{0.9}

\usepackage[T1]{fontenc}

\usepackage[utf8]{inputenc}

\usepackage{microtype}

\usepackage{inconsolata}

\usepackage{graphicx}

\usepackage{subcaption}
\usepackage{booktabs}
\usepackage{amsmath}
\usepackage{graphicx} 
\usepackage[utf8]{inputenc} 
\usepackage[T1]{fontenc}    
\usepackage{hyperref}       
\usepackage{url}            
\usepackage{booktabs}       
\usepackage{amsfonts}       
\usepackage{nicefrac}       
\usepackage{microtype}      
\usepackage{xcolor}         
\usepackage{makecell}

\newcommand{\NB}[1]{{\color{purple} (Neil: #1)}}
\renewcommand{\NB}[1]{} 
\newcommand{\irena}[1]{{\color{cyan} (Irena: #1)}}
\renewcommand{\irena}[1]{}

\title{Benchmarking Distributional Alignment of Large Language Models}

\author{%
  Nicole Meister \\
  Stanford University\\
  \texttt{nmeist@stanford.edu} \\
  \And
  Carlos Guestrin \\
  Stanford University\\
  \texttt{guestrin@stanford.edu} \\
  \And
  Tatsunori Hashimoto \\
  Stanford University\\
  \texttt{thashim@stanford.edu} \\
}

\begin{document}

\maketitle

\begin{abstract}
\input{Sections/abstract}
\end{abstract}

\section{Introduction}

\input{Sections/intro}

\section{Problem Statement}
\input{Sections/problemsetting}

\section{Benchmark Construction}
\input{Sections/benchmark_acl}

\section{Experiments}
\input{Sections/experiments_acl}

\section{Discussion}
\input{Sections/discussion}

\section{Related Work}
\input{Sections/relatedwork}

\section{Conclusion}
\input{Sections/conclusion}

\clearpage

\section{Limitations}
\input{Sections/limitations}

\section{Ethical Considerations}
\input{Sections/ethical_considerations}

\clearpage

\bibliography{custom}

\clearpage

\appendix
\input{Sections/appendix_acl}

\end{document}

%% file: Sections/abstract.tex
Language models (LMs) are increasingly used as simulacra for people, yet their ability to match the distribution of views of a specific demographic group and be \textit{distributionally aligned} remains uncertain.
This notion of distributional alignment is complex, as there is significant variation in the types of attributes that are simulated.
Prior works have underexplored the role of three critical variables---the question domain, steering method, and distribution expression method---which motivates our contribution of a benchmark explicitly addressing these dimensions.
We construct a dataset expanding beyond political values, create human baselines for this task, and evaluate the extent to which an LM can align with a particular group's opinion distribution to inform design choices of such simulation systems. 
Our analysis reveals open problems regarding if, and how, LMs can be used to simulate humans, and that LLMs can more accurately describe the opinion distribution than simulate such distributions. 



%% file: Sections/intro.tex

It would be unusual to ask a person to accurately simulate a demographic group to which they do not belong. However, LMs are increasingly being used in this way to simulate human behavior in applications ranging from agent-based simulations \citep{park2023generative} to piloting survey design \citep{hwang2023aligning, zhou2024sotopia, pmlr-v202-aher23a, ziems2024large,Argyle_2023}. 
When simulating survey responses, there is no single ``correct'' answer, and it is important to evaluate if the \textit{distribution} of model outputs is truly aligned with the intended human distribution. There has been considerable debate as to whether or not models can do this---some argue that the extensive training corpus of LLMs enables them to faithfully simulate demographic groups \citep{grossman2023}, while others show such simulations are inaccurate and stereotypical~\citep{liu2024evaluating,wang2024large}. 

One reason for these conflicting views is the heterogeneity in how one can measure distributional alignment, resulting in a lack of clarity around best practices. 
For example, current approaches have measured the model's opinion distribution with zero-shot, log-probability based evaluations, yet recent work in uncertainty quantification suggests verbalized distributions can outperform model log-probabilities \citep{tian2023justaskcalibrationstrategies}. 
This raises the question of whether the model's distribution expression method is truly optimal and demonstrates a need for carefully controlled evaluations.




\begin{figure*}[ht!]
  \centering
  \includegraphics[width=\textwidth]{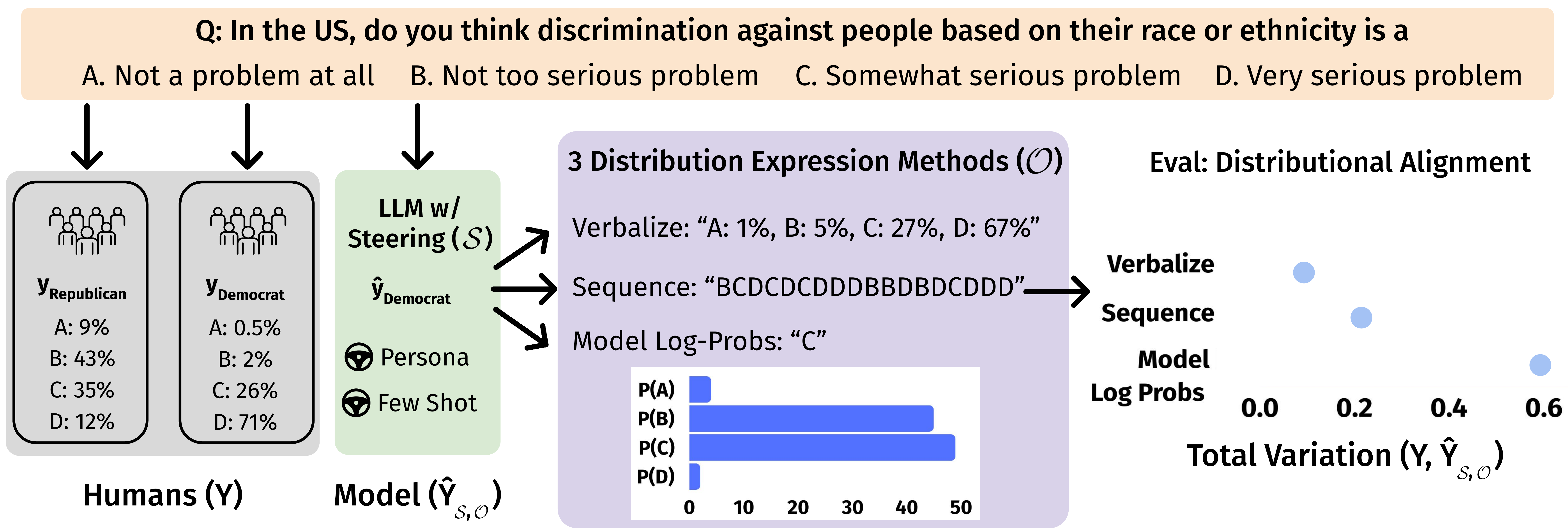}
  \caption{
  Our work studies how variations in the dataset (yellow), steering method (green), and distributional expression method (purple)
  affect the quality of distributional alignment. We rank models and humans in their ability to align with the opinion distribution of demographic groups and find existing metrics for distributional alignment (i.e., model log-probabilities) systematically underestimate LM performance. While LMs may `know' about distributional alignment, they struggle to sample from their own distribution. 
  }
\label{fig:fig1}
\end{figure*}


In this work, we acknowledge the sensitivity of distributional alignment metrics and build a benchmark that studies several key variations in the distributional alignment task. Our benchmarks and dataset measure how the distributional alignment of LLMs vary under (1) the distribution expression method, (2) the steering method, and (3) design choices in the dataset (Fig.~\ref{fig:fig1}).

Our analyses reveal several open problems for distributional alignment. First, we find that existing measurement methods such as log-probabilities have systematically underestimated the distributional alignment of LLMs, and other simple baselines result in better alignment. Second, we find that LMs can more accurately estimate opinion distributions in text-based forms (e.g., `return the distribution in a JSON'), compared to generating samples from the opinion distribution. This highlights a substantial opportunity to improve distributional alignment by closing the gap between a model's knowledge of human opinions and its ability to simulate them. Finally, we find significant gaps in both alignment and steerability when simulating non-cultural opinions, such as book preferences, when compared to evaluations of stronger opinions (i.e., political and cultural values). 






We summarize our key contributions as follows:
\begin{enumerate}
    \item We identify three key sources of variation in distributional alignment (the question, steering method, and distribution estimation method) and construct a benchmark systematically varying these dimensions.
    \item We collect a new dataset, NYT Book Opinions, that expands measurements beyond political and cultural values. 
    \item Our analysis reveals several open problems for the field: (1) LMs may `know' a distribution, but are unable to sample from it (2) Log-probability-based metrics for distributional alignment may systematically underestimate LM performance (3) Distributional alignment and steering beyond political and cultural values remains challenging.
\end{enumerate}

%% file: Sections/problemsetting.tex
\label{sec:problemsetting}
We propose a benchmark that systematically evaluates the extent to which a language model can be aligned to the distribution of a particular demographic group's opinions, a task we term the \textit{distributional alignment problem}.
To begin, we formalize this task and visualize it in Fig.~\ref{fig:fig1}.  


Let $q \in Q$ be a survey question to which respondents from group $g \in G$ have an opinion distribution across multiple choices answers $y_{g,q}$. 
The goal is to understand how a language model can represent a group $g$ through a steering method $S$, one that shifts an LM's opinion distribution to that of a particular group. 
Concretely, the model will express a distribution $\hat{y}_{g,q}$ with a distribution expression method $\mathcal{O}$ (e.g., model log-probabilities). 

We are interested in the distributional difference between the reference distribution, $y_{g,q}$, and the model's estimate, $\hat{y}_{g,q}$. To evaluate this, 
we construct a set\footnote{This is not a matrix as each question $q$ can have a different number of answer choices. Thus, the dimensionality of $y_{g,q}$ depends on $q$.} $Y$ of ground truth human opinion distributions, where 
$Y = \{ y_{g,q} \mid 1 \leq g \leq G, 1 \leq q \leq Q \}$, and a corresponding set $\hat{Y}_{S, \mathcal{O}}$ of a model's predicted distributions, where
$\hat{Y}_{S, \mathcal{O}} = \{ \hat{y}_{g, q} \mid 1 \leq g \leq G, 1 \leq q \leq Q \}$. We define \textbf{distributional alignment} as 
\begin{equation}
\label{eq:distributional_alignment}
\mathcal{A}(Y,\hat{Y}_{S, \mathcal{O}}) = \frac{1}{|G|} \sum_{g \in G}  \frac{1}{|Q|} \sum_{q \in Q} \frac{1}{2} {\left \lvert\lvert y_{g, q}- \hat{y}_{g, q} \right \rvert\rvert}_1 .
\end{equation}
This metric is the average total variation between these two distributions, with a smaller number representing a higher 
performance on the task.

%% file: Sections/benchmark_acl.tex
\label{sec:benchmark}

Having formalized the notation for distributional alignment, we explore how it can be improved by focusing on three understudied sources of variation: the distributional expression method ($\mathcal{O}$), steering method ($S$), and dataset ($Y$).
In this section, we explain how these elements are used to construct the benchmark and describe the human baseline.
\subsection{Distributional Estimation Method ($\mathcal{O}$)}
\label{sec:benchmark-estim-method}


\begin{figure*}[ht!]
  \centering
  \includegraphics[width=0.9\textwidth]{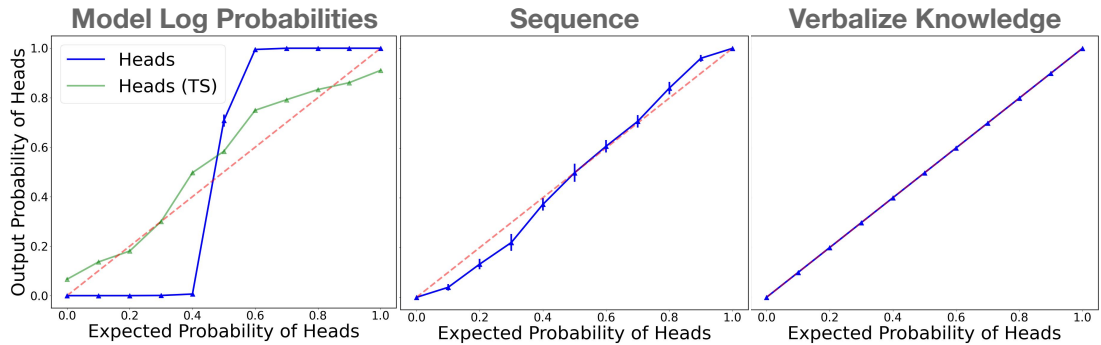}
  \caption{\textbf{Biased Coin Flip}: We find that when the probability of heads is measured via model log-probabilities (left), the results are highly uncalibrated (this behavior is mitigated with temperature scaling (TS), shown in green). However, when the distribution is expressed through emitting a 30-token sequence of H or T (Sequence) or directly verbalizing the distributional knowledge (Verbalize Knowledge), we do not observe the same mis-calibration.}
\label{fig:biasedcoin}
\end{figure*}


In this section, we describe three distributional expression methods and demonstrate how distributional alignment is highly sensitive to the distribution estimation method.\footnote{Full prompts in \href{https://github.com/nicolemeister/benchmarking-distributional-alignment}{github.com/nicolemeister/benchmarking-distributional-alignment.}}



\textbf{1. Model Log-probabilities: }
A model's next token log probabilities assigned to each of the answer choices (e.g., ‘A’, ‘B’...) provide a sampling distribution by directly sampling from the model. This is the canonical distribution estimation method \citep{santurkar2023opinions, durmus2024measuring};
however, prior work has found that these model log-probabilities have a concentrated probability mass on a few answer choices rather than more dispersed answers as seen in their corresponding human distributions \citep{durmus2024measuring}, especially in models trained with Reinforcement Learning from Human
Feedback (RLHF) \citep{christiano2023deepreinforcementlearninghuman}.

\textbf{2. Sequence of Tokens: }While model log-probabilities represent drawing samples from a LM,
a model can also express its distributions by behaving as a sampler. We instruct a model to emit a sequence of 30 samples from the distribution (e.g., \texttt{ABBBAABDDBACBDB}). 
This method is advantageous when practitioners want to generate samples from an opinion distribution for simulation purposes.\footnote{This is inspired by a common synthetic data generation prompting technique that instructs LLMs to emit a sequence or batch of answers to generate diverse samples \citep{wang2023selfinstructaligninglanguagemodels,dubois2024alpacafarmsimulationframeworkmethods,si2024llmsgeneratenovelresearch}.} 
This distribution is limited by the number of tokens emitted in the sequence, as we are trying to estimate a continuous distribution from a finite number of tokens. 
Thus, we report a \textbf{discretization error}---the error incurred when drawing 30 samples from the ground truth distribution and computing the total variation based on those samples.


\textbf{3. Verbalize Distributional Knowledge: } 
Lastly, we can remove the requirement that models must sample from simulated humans and instruct a model to directly verbalize the distribution in the output text, without an estimation procedure or post-processing applied (e.g., text in a JSON format \texttt{\{A: 25\%, B: 20\%, C: 45\%, D: 10\%\}}).\footnote{While this verbalized knowledge can be fed into an external random sampler, it is not a sufficient output for downstream applications (e.g., piloting surveys) and existing approaches to simulating humans do not take this approach \cite{park2023generative,samuel2024personagymevaluatingpersonaagents}. Our later results suggest this approach may be fruitful as a distributional alignment method.}
This differs from the aforementioned methods in that it separates distributional knowledge from the LM's ability to also generate samples.

\textbf{These three estimation methods reveal surprising performance gaps.} Consider a toy experiment in which the model has \textit{full knowledge} of the ground truth distribution.
In this experiment, an LLM is instructed to simulate a flip of a biased coin which has $P(H)=p$ and $P(T)=1-p$. Naturally, we would expect the model log-probabilities for the token `H' to be $p$ and `T' to be $1-p$. In reality, we see that the model log-probabilities are highly un-calibrated. They provide a misleading picture of the ground truth distribution, despite the fact that the ground truth distribution is shared in the input prompt (Fig.~\ref{fig:biasedcoin}). Moreover, we find that two other distributional estimation methods do not suffer the same mis-calibration issue: both verbalizing the distribution and emitting a sequence of 30 samples of the biased coin flip are much more calibrated than the model's log-probabilities.\footnote{This performance gap is not unexpected; it has been shown that while models excel in text and image generation tasks (knowledge), they fall short when asked to validate if the generated answer is correct \citep{li2023benchmarking,west2023generative}.}



Results from this biased coin flip experiment suggest two things: (1) prior distributional alignment work using model log-probabilities may not be seeing the full picture and (2) there exists a significant performance gap, one that we term \textit{knowledge-to-simulation gap}. This gap refers to instances where models may have alignment in knowledge (i.e., verbalizing the distribution in the output text), but not in sampling behavior (i.e., as measured via model log-probabilities or emitting a sequence of tokens). 

We define the \textbf{knowledge-to-simulation gap} as the percent error between the alignment when emitting a sequence of answer choices and verbalizing the distribution.\footnote{The gap can also defined as the difference in performance between verbalize and model log-probabilities, yet our later results show model log-probabilities to not be competitive.} More formally, this gap is: 
\begin{equation}
\label{eq:knowledge_to_sim}
    \text{KS}_S = \frac{\mathcal{A}(Y,\hat{Y}_{S, {\text{Sequence}}})}{\mathcal{A}(Y,\hat{Y}_{S, {\text{Verbalize}}})}-1.
\end{equation}

\subsection{Steering Method ($S$)}

\label{sec:benchmark-steeringmethod}

Steerability in the context of this work refers to a LM’s ability to adapt to represent the opinion of a target demographic group.
We evaluate two steering methods, persona and few-shot steering, by prepending additional context to the prompt describing the group we want the model to emulate.

We chose to study this few-shot setting, as it is known that persona steering can be inaccurate, leading to undesirable side effects including stereotyping, exacerbating polarization, and creating echo chambers \citep{perez2022discovering, cheng2023marked, wang2024large}.

\textbf{Persona Steering}: 
\citet{cheng2023marked} define a persona as a ``natural language portrayal of an imagined individual belonging to some (intersectional) demographic group.'' In persona steering, we append a persona to the prompt and ask the LM to emulate behavior from this group. Concretely, we follow a version of persona steering from \citet{santurkar2023opinions,Kambhatla} where the LM is instructed to pretend to be a member of the target demographic group (e.g. ``Please simulate an answer from a group of Democrats.''). 

\textbf{Few Shot Steering}: 
Inspired by the success of few-shot prompting in language understanding tasks \citep{NEURIPS2020_1457c0d6}, 
we construct a few shot setting in which in-context examples of ground truth group opinion distributions are provided in addition to the persona. Specifically, LMs are given the top five most similar questions and their corresponding ground truth distribution from a group, and instructed to simulate an answer from that group (see Appendix~\ref{sec:appendix-fewshot} for more details). This setting is representative of when practitioners have access to existing survey data for similar questions.


\textbf{No Steering}: We can directly contrast steered models to models that are prompted with a question without any demographic or identity markers. 



\subsection{Dataset ($Y$)}
\label{sec:benchmark-datasets}

In this section, we describe two complementary datasets for quantifying distributional alignment -- OpinionQA~\citep{santurkar2023opinions} and a new non-political subjective opinion dataset, NYT Book Opinions.



\textbf{OpinionQA: }We use the OpinionQA dataset from \citet{santurkar2023opinions} to leverage public opinion surveys to compare the distribution of LLM responses to those of US citizens. In their steerability analysis, they create a smaller set of 500 contentious questions where the subgroups frequently disagree. We follow suit and randomly sample 100 questions from this set to obtain questions spanning topics such as science, politics, and personal relationships. We obtain the ground truth human opinion distributions of PEW survey respondents belonging to six demographic groups: Democrat, Republican, Male, Female, Black, and White.


\textbf{NYT Book Opinions: } 
There are several works studying how LMs respond to political opinions or cultural values \citep{santurkar2023opinions,durmus2024measuring}, but it is less understood how LMs respond to non-political, yet still subjective values. 
How do our findings extend to other domains of personalization? Are LMs still suitable for this use case? 

This motivates the construction of a new dataset, \textbf{NYT Book Opinions}, that gathers opinions on interest in the top books from the past two decades as judged by \citet{nytimes2024topbooks}. The purpose is to capture subjective values that less directly measure cultural values and political leanings.

\textbf{Annotation setup: } We collected 235 books and their corresponding author, book summary, and genre. 346 annotators provided a 4-point Likert rating to the question, ``Given the summary of this book, how likely are you to read it?'' for 26 books. See Appendix for additional details.

\begin{figure}[h!]
  \centering
  \includegraphics[width=\linewidth]{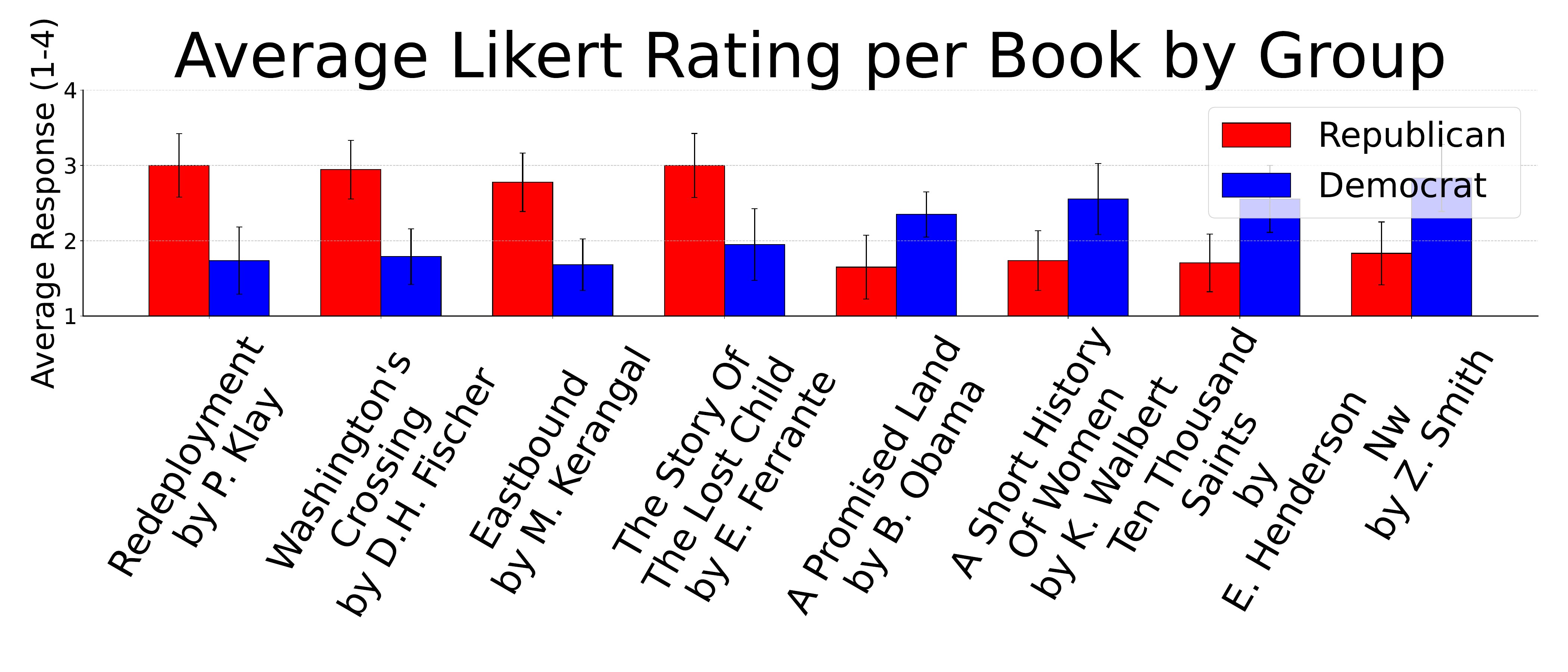}
  \caption{Top 4 books with the largest difference between Republican and Democrat ratings (left) and the top 4 books with the largest difference between Democrat and Republican ratings (right), with a 95\% confidence interval from bootstrapping. }
\label{fig:nytimes}
\end{figure}

From these human annotations, we find disagreement in book interest. In Fig.~\ref{fig:nytimes}, we plot the top 4 books that have the largest difference in Democrat and Republican annotator ratings.
Books that Republican annotators preferred over Democrat annotators include works such as \textit{Redeployment} by Phil Klay (twelve stories by a former Marine who served in Iraq) and \textit{Washington's Crossing} by David Hackett Fischer (a story from the American Revolutionary War). Books that Democrat annotators preferred over Republican annotators include works such as \textit{A Promised Land} by Barack Obama and 
\textit{A Short History Of Women} by Kate Walbert.

\subsection{Human Baseline Annotations}
    \label{sec:benchmark-human}
Inspired by~\citet{yudkin_hawkins_dixon_2019} who study the \textit{Perception Gap}, or the percentage difference between a respondent’s estimate of how many people held a certain view and the actual percentage of people who held that view, we recruit crowd workers to complete the distributional alignment task. The annotators receive the same questions from OpinionQA and NYT Book Opinions that we evaluated models on, allowing us to compare human performance against the suite of LMs we evaluate. As with the models, the human annotators experience three steering methods: no steering, persona steering, and few-shot steering. Each survey question or book summary receives four annotations of estimated opinion distributions over answer choices.

%% file: Sections/experiments_acl.tex
\label{sec:experiments}

We rank GPT-4, GPT-3.5, Claude Haiku, Claude Opus, Llama-3 70B Instruct,\footnote{Smaller models struggled to follow the sequence distribution expression method, thus restricting our model selection.} and humans based on distributional alignment (Eq.~\ref{eq:distributional_alignment}) and the knowledge-to-simulation gap (Eq.~\ref{eq:knowledge_to_sim}), and average across groups, steering methods, and datasets. 
We start by describing the performance on the distributional alignment task.
Then, we dive into the implications that emerge from varying the distribution expression method, dataset, and steering method. 
\NB{I feel like you start experiments with the key questions you want to answer and also maybe some headline results that you find. if you end up going with the more active pitch of "verbalized distributions perform best" you could mention this here, I think}




\subsection{Distributional Alignment Performance}
\label{sec:exp-performance}

\begin{table}[ht!]
  \centering
  \begin{minipage}{\linewidth}
    \centering
    \begin{tabular}{ll}
      \toprule
      Model     & $\mathcal{A}(Y,\hat{Y}_{\mathcal{S}, \mathcal{O}})$  \\
      \midrule
            \rowcolor{lightgray}
GPT-4 (V) & 0.204 $\pm$ 0.004 \\
      \rowcolor{lightgray}
Anthropic Opus (V) & 0.219 $\pm$ 0.005 \\
      \rowcolor{lightgray}
Llama 3 70B (V) & 0.226 $\pm$ 0.004 \\
      \rowcolor{lightgray}
Anthropic Haiku (V) & 0.235 $\pm$ 0.005 \\
GPT-4 (Seq) & 0.237 $\pm$ 0.004 \\
      \rowcolor{lightgray}
      Humans (V) & 0.247 $\pm$ 0.004 \\
      \rowcolor{lightgray}
GPT-3.5-Turbo (V) & 0.259 $\pm$ 0.005 \\
GPT-4 (TS-Log-p) & 0.260 $\pm$ 0.004 \\
GPT-3.5-Turbo (Seq) & 0.278 $\pm$ 0.005 \\
Anthropic Haiku (Seq) & 0.287 $\pm$ 0.006 \\
GPT-3.5-Turbo (TS-Log-p) & 0.290 $\pm$ 0.005 \\
Llama 3 70B (Seq) & 0.320 $\pm$ 0.006 \\
Anthropic Opus (Seq) & 0.337 $\pm$ 0.006 \\
GPT-3.5-Turbo (Log-p) & 0.462 $\pm$ 0.007 \\
Llama 3 70B (TS-Log-p) & 0.460 $\pm$ 0.007 \\
Llama 3 70B (Log-p) & 0.515 $\pm$ 0.006 \\
GPT-4 (Log-p) & 0.582 $\pm$ 0.006 \\
            \midrule
            Discretization Error (Seq) & 0.126 $\pm$ 0.006  \\
            Uniform & 0.302 $\pm$ 0.005   \\
      \bottomrule
    \end{tabular}
    \subcaption{\textbf{Distributional Alignment Task}. Models ranked based on mean total variation. Models highlighted in gray and with (V) have a distribution expression method of directly verbalizing the distribution in a JSON format ($\mathcal{O} = \texttt{Verbalize}$). Models not highlighted represent samplers, where (Seq) represents the 30-token sequential distribution output ($\mathcal{O} = \texttt{Sequence}$), (Log-p) represents $\mathcal{O} = \texttt{Model Log-probabilities}$, and (TS-Log-p) represents $\mathcal{O} = \texttt{Temperature Scaled Model Log-probabilities}$. }
    \label{tab:main-a}  
    

  \end{minipage}%
  \hspace{0.01\textwidth} 
  \begin{minipage}{\linewidth}
    \centering
    
    \begin{tabular}{lc}
      \toprule
      Model     & Simulation Penalty     \\
      \midrule
      GPT-3.5 Turbo & 8.39\%  \\
      GPT-4 & 19.23\%  \\
      Claude: Haiku & 24.64\%   \\
      Llama 3 70B Instruct & 41.86\%    \\
      Claude: Opus & 53.57\% \\
      \bottomrule
    \end{tabular}
    
    \subcaption{\textbf{Knowledge-to-Simulation Gap (Eq.~\ref{eq:knowledge_to_sim})}. The simulation penalty measures the percent error increase in total variation between the 30-token sequential distribution output and the verbalization of knowledge.}
    \label{tab:main-b}
  \end{minipage}
\caption{\textbf{Model and Human Performance}. In both tables, we average over the OpinionQA and NYT dataset, persona and few shot steering, and all demographic groups. We report the 95\% confidence interval from bootstrapping with 1000 samples and rank models from highest to lowest performance on the task.}
\end{table}

In Tab.~\ref{tab:main-a}, we report the results of our distributional alignment leaderboard where we rank models and humans on their ability to be steered towards a demographic group, averaged over persona steering and few shot steering, and both datasets. In this leaderboard, where lower numbers represent higher distributional alignment, we find that verbalizing the distribution results in higher performance, with GPT-4 being the most steerable amongst our models, with an average total variation of 0.204, followed by Opus (0.219) and then Llama-3-70B (0.225). These numbers can be directly compared to the performance of the uniform baseline, where each answer option is equally likely to occur in the sequence (0.382), and human annotations that achieve an average total variation of 0.247.

\subsection{Implications for Distributional Alignment}
\label{sec:exp-implications}
In this section, we organize our analyses into implications for the field and conclude each section with actionable suggestions for practitioners who use LLMs for simulating human subjects.


\textbf{A large knowledge-to-simulation gap exists. }As observed in Sec.~\ref{sec:benchmark-estim-method}, even when a model knows as distribution, sometimes it cannot sample it. To this end, we measure these gaps between knowledge and simulation in our second leaderboard (Tab.~\ref{tab:main-b}). We observe that some models, particularly those from the Llama-3 and Claude suite have a larger knowledge-to-simulation gap, especially when directly compared to the OpenAI models. This highlights room for improvement, as the language model is capable of returning more accurate estimates of human opinions when expressing the distribution through verbalizing the distribution, but not when simulating samples from this distribution.
\textbf{\textit{Implication: }}If practitioners are using models with a high knowledge-to-simulation gap, they should prompt the model to verbalize the knowledge directly and use an alternative sampler for simulation purposes.
\NB{shouldn't the implication be: if you want the distribution, use verbalized; if you want samples, still used verbalized, but use an external sampler?}


\textbf{Model log-probabilities are misleading.} Next, we highlight a larger knowledge-to-simulation gap between model log-probabilities and the verbalization of knowledge (Tab.~\ref{tab:main-a}). Using model log-probabilities to measure distributional alignment results in worse performance than even \textit{uniform} distribution on the distributional alignment task. We find that this performance gap can be in part attributed to log probabilities having a highly concentrated probability mass on one or two answer choices, as observed in \citet{durmus2024measuring} and our biased coin flip experiment from Sec.~\ref{sec:benchmark-estim-method}.
While prior work \citep{santurkar2023opinions,durmus2024measuring} has used model log-probabilities for steering and alignment applications, this distribution expression method \textit{underestimates} LLM capabilities in emulating demographic groups. 
\textbf{\textit{Implication: }} We encourage practitioners to use alternative distribution estimation methods, such as emitting a sequence or verbalizing the distributional knowledge.

\begin{figure}[t!]
  \centering
  \includegraphics[width=\linewidth]{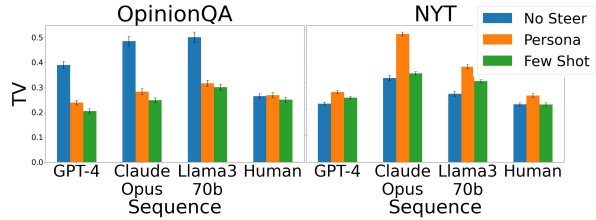}
  \caption{\textbf{Steering Method and Dataset: } We plot the average total variation for each dataset and steering method, averaged across demographic groups for the 30-token sequential distribution output. We find it is harder to steer models toward the dataset where opinions are hidden under a layer of abstraction (NYT). Additionally, few shot steering improves distributional alignment for humans and all models except for GPT-3.5.  \NB{could you maybe make some other strong claims based on these results? in particular it looks like GPT-4 is competitive or better than the human baseline -- that's pretty interesting/notable/a kinda flashy headline result?}}
  \label{fig:OQAvsNYT}
\end{figure}

\textbf{Steering is more challenging in non-cultural and non-political settings. }
The goal of the NYT Book Opinions dataset is to capture subjective values from demographic groups that less directly measure cultural values and political leanings, unlike OpinionQA questions which directly ask value-laden survey questions. We find that all models and humans are more easily steered towards questions in the OpinionQA dataset than the NYT Book Opinions dataset, as the total variation decreases from no steering to persona and few shot steering.
In Fig.~\ref{fig:OQAvsNYT}, we plot the average total variation for each dataset and steering method, averaged across demographic groups, for the output type of sequence. 
This suggests that when the values are hidden under a layer of abstraction (i.e., book interest) it may be harder to steer models towards the opinions of demographic groups. 
\textbf{\textit{Implication:} }When practitioners use LLMs to pilot survey questions that less directly allude to cultural values, they need to consider the nature of their questions and if LLMs are suitable for their use case. 

\textbf{Few shot steering improves persona steering. }   
For humans and all models except GPT-3.5, we observe statistically significant improvements in few shot steering over persona steering (Fig.~\ref{fig:OQAvsNYT}). As expected, when models have access to five examples of distributional data, they improve their distributional estimation capabilities.
\textbf{\textit{Implication: }}Practitioners should use prior distributional data as few shot examples over just personas, if possible.

\begin{figure}[t!]
  \centering
  \includegraphics[width=\linewidth]{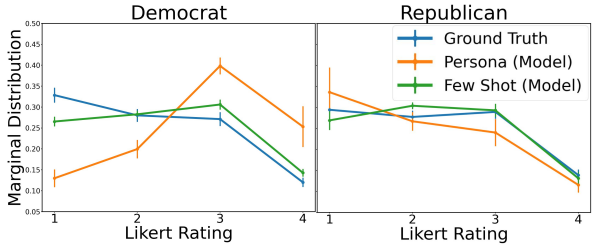}
  \caption{\textbf{Models assume Democrats read more than Republicans.} In this plot, we show the marginal distribution of Likert Rating (1-4) in responses to the following question: ``How likely are you to read this book?'' A Likert rating of 1 refers to ``Very unlikely'' and a Likert rating of 4 refers to ``Very likely''. 
  We averaged over 235 questions from NYT Book Opinions and 5 models steered towards Democrats and Republicans with persona steering (orange) and few shot steering (green). In blue, we plot the reference human reference for Democrat and Republican annotators. We find that persona steering produces more stereotypical results.}
  \label{fig:NYT_stereotyping}
\end{figure}

\textbf{Persona-steered LLMs are susceptible to stereotyping. }Consistent with prior work studying persona steering and LLMs emulating human behavior \citep{gupta2024biasrunsdeepimplicit,cheng2023marked,cheng2023compostcharacterizingevaluatingcaricature,wan2023personalizedstochasticparrotsdangerous}, we find that the LLMs produce stereotypical outputs. For example, when analyzing the marginal distribution of the answer ratings in the NYT Book Opinions dataset (Fig.~\ref{fig:NYT_stereotyping}), we find that models prompted with a Democrat persona tend to simulate humans that are more likely to read books---the average simulated human has a 13\% chance of responding with ``very unlikely to read'' when our human annotators had a 33\% chance. Furthermore, the simulated Democrat has a 25\% chance of responding with ``very likely to read'', when our human annotators had a 12\% chance. This gap is significantly reduced when using few-shot steering. \textbf{\textit{Implication: }}It is important to collect disaggregated evaluation metrics to understand potential discrepancies between groups \citep{barocas2021designingdisaggregatedevaluationsai}. These findings support using prior distributional data as a few shot examples.


%% file: Sections/discussion.tex
\label{sec:discussion}

\subsection{Human Performance}

\begin{table}[ht!]
  \centering
  \begin{minipage}{\linewidth}
    \centering
    \begin{tabular}{ll}
      \toprule
      Model     & $\mathcal{A}(Y,\hat{Y}_{\mathcal{S}, \mathcal{O}})$  \\
      \midrule
GPT-4 (V) & 0.204 $\pm$ 0.003 \\
Anthropic Opus (V) & 0.219 $\pm$ 0.004 \\
Llama 3 70B (V) & 0.225 $\pm$ 0.004 \\
Anthropic Haiku (V) & 0.235 $\pm$ 0.004 \\
GPT-4 (Seq) & 0.237 $\pm$ 0.004 \\
      Humans (V) & 0.250 $\pm$ 0.004 \\
GPT-3.5-Turbo (V) & 0.259 $\pm$ 0.005 \\
      \bottomrule
    \end{tabular}


  \end{minipage}%
  \hspace{0.01\textwidth} 
     
\caption{\textbf{Distributional Alignment Task with Human Performance (OQA, NYT)}. Models with a distribution expression method of directly verbalizing the distribution (V) are ranked based on mean total variation. We average over the OQA and NYT, persona and few shot steering, and all demographic groups, and report the 95\% CI from bootstrapping with 1000 samples. For humans, we compute the average over annotators per question and report the 95\% CI from bootstrapping with 1000 samples over questions.}
 \label{tab:human}  
\end{table}

In Tab.~\ref{tab:human}, we contextualize the performance of LMs with human annotators who attempt to guess the opinions of others. Although the best LMs with the most effective distribution methods (verbalize) perform close to this human baseline, this is not particularly promising for the field of distributional alignment given that humans are known to be poor predictors of opinions of the opposite party~\citep{yudkin_hawkins_dixon_2019,levendusky}. It would be highly questionable to base the result of social science surveys on participants guessing others' opinions, and our findings indicate that LMs offer little improvement over this baseline.

\subsection{Open Problems} Our analyses reveal unique challenges for the community to make progress on. In this section, we lay out those open problems.

\textbf{Knowledge-to-Simulation Gap.} First, we find in the biased coin flip experiment, and then more rigorously with survey data (Tab.~\ref{tab:main-b}), that while LMs may `know’ a distribution, they struggle to sample from their own distribution. Future work should address the sampling capabilities of models and why they struggle with randomness and representing distributions (e.g., \citet{requeima2024llmprocessesnumericalpredictive} and \citet{paruchuri2024oddslanguagemodelscapable}). 

\textbf{Misleading Model Log-probabilities.}
Existing research has considered distributional alignment through model log-probabilities; however, our benchmark reveals this method's shortcomings, as models evaluated by log-probabilities fail to rank among the top ten in our distributional alignment leaderboard.
Future work should address why model log-probabilities are mis-calibrated in distributional alignment settings and pivot to improving sampling through emitting a sequence of tokens.

\textbf{Limitations of Persona Steering.}
Few shot steering improves the performance of persona steering, suggesting that models lack key information about the opinions that a few examples can provide. We find this is often due to persona-steered models conceptualizing humans as less nuanced and more polarized. This reveals a clear challenge in building models that can capture the idiosyncracies of a person and avoid extremized stereotypes.



%% file: Sections/relatedwork.tex
\label{sec:relatedwork}


\textbf{Distributionally Pluralistic Alignment. }
LLMs often learn an averaged human preference and struggle to model diverse preferences across groups. Recent works advocate for \textit{distributionally pluralistic models} that are well-calibrated to a group's distribution of responses \citep{sorensen2024roadmap,feng2024modularpluralismpluralisticalignment,kirk2024prismalignmentprojectparticipatory, chen2024palpluralisticalignmentframework}. However, \citet{sorensen2024roadmap} acknowledge there is limited knowledge of explicit alignment procedures to increase distributional calibration, highlighting the 
importance of our work in characterizing key sources of variation and how they affect distributional alignment.

\textbf{LLMs for Simulating Human Behavior.}
With the proliferation of LLMs, recent work has integrated LLMs into computational social science to simulate social psychology experiments \citep{pmlr-v202-aher23a,DILLION2023597}, create human-like agents \citep{park2023generative,samuel2024personagymevaluatingpersonaagents,horton2023largelanguagemodelssimulated}, and annotate data 
\citep{he2024annollmmakinglargelanguage,doi:10.1177/20531680241231468} to name a few. We focus on a popular use case of LLMs simulating humans to generate survey samples \citep{hwang2023aligning, zhou2024sotopia, pmlr-v202-aher23a,Argyle_2023}.

Several works urge caution when relying on the survey responses of LLMs to elicit synthetic responses, citing concerns such as group stereotyping and misrepresentation \citep{wang2024large,abdurahman_perils,geng2024largelanguagemodelschameleons}, preference for socially desirable responses \citep{ai2024cognitionactionconsistentnot}, 
lower entropy in model responses \citep{dominguezolmedo2024questioningsurveyresponseslarge,park2023diminisheddiversityofthoughtstandardlarge}, and answer inconsistencies from prompt brittleness \citep{ceron2024promptbrittlenessevaluatingreliability}.
While these works provide important context, they focus on zero-shot and political or cultural values, leaving several sources of variation unexamined. The closest work to ours is \citet{dominguezolmedo2024questioningsurveyresponseslarge} who study answer choice order bias and find that model responses have different variation than that of humans. Our work is distinct in that we look beyond stability to prompt variation, and focus on higher-level design choices such as the steering method (e.g. few shot) and distribution estimation method, which have a significant impact on alignment measurements.


Next, we describe existing research on these variables and how our work makes new contributions.

\textbf{Dataset.}
\citet{santurkar2023opinions}
quantify alignment through responses to PEW surveys, inspiring numerous works \citep{durmus2024measuring,naous2024havingbeerprayermeasuring,wang2024countriescelebratethanksgivingcultural,pistilli2024civicsbuildingdatasetexamining,kovač2023largelanguagemodelssuperpositions,masoud2024culturalalignmentlargelanguage,zhao2024worldvaluesbenchlargescalebenchmarkdataset,stammbach2024aligninglargelanguagemodels,röttger2024politicalcompassspinningarrow}. 
However, there is no publicly available dataset on distributional preferences to non-political yet subjective values (e.g., product preferences)
motivating our NYT Book Opinions dataset.

\textbf{Steering Method.}
The literature has studied a variety of methods to steer the generation of LLMs toward specific opinions. A popular method of steering is persona steering, achieved by prepending demographic information to prompts \citep{santurkar2023opinions,simmons2023moralmimicrylargelanguage,perez2022discovering,cheng2023marked},
prepending past opinions \citep{hwang2023aligning}, or fine-tuning \citep{jiang2022communitylmprobingpartisanworldviews,namikoshi2024usingllmsmodelbeliefs}. 
Prior works have also explored steering with in-context examples \citep{kim2024fewshotpersonalizationllmsmisaligned,zhao2023grouppreferenceoptimizationfewshot}, prefix-tuning with persona grounded in collaborative filtering \citep{li2024steerabilitylargelanguagemodels}, modifying activations in the forward pass \citep{turner2024activationadditionsteeringlanguage}, and creating human belief networks \citep{chuang2024demographicsaligningroleplayingllmbased}, yet none have studied the differences between persona and few shot steering as we do. 
Closest to our work is 
\citet{santurkar2023opinions} and \citet{liu2024evaluating}, who evaluate persona steering, yet we are unique in that we compare persona steering to instances where practitioners have access to prior survey data and can use it as few shot examples. 

\textbf{Distribution Expression Method.} 
Prior work has found that prompting LMs for statement
probabilities (e.g., verbalize) and model log-probabilities can sometimes lead to different results~\citep{hu2023prompting,liu-etal-2023-cognitive}.
When evaluating LLMs on the basis of survey questions, \citet{santurkar2023opinions}, and all the works that have followed, study models’ log-probability distribution over various answer choices. Recent work has begun to explore randomness in coin flips \citep{vankoevering2024randomrandomevaluatingrandomness,mondal2024largelanguagemodelsexhibit}, improved sampling from LMs \citep{zhu2024recoveringmentalrepresentationslarge,requeima2024llmprocessesnumericalpredictive} and using LMs for probabilistic reasoning \citep{paruchuri2024oddslanguagemodelscapable}, but do not apply it to the simulating opinion distributions. 
We explore differences in distribution expression and take inspiration from work that suggests verbalized uncertainty can be competitive with model log-probabilities \citep{tian2023justaskcalibrationstrategies,mondal2024largelanguagemodelsexhibit}.

%% file: Sections/conclusion.tex
LLMs perform surprisingly well on knowledge-intensive tasks, excelling on coding benchmarks and question-answer tasks to name a few. 
Their success in these tasks has led to an increase in applying LLMs to simulate human behavior, yet their ability to accurately reflect specific demographic groups remains controversial. 
To study this problem, we construct a benchmark to rank humans and models by performance on the distributional alignment task. Our findings reveal many unresolved challenges in distributional alignment, notably the model's sensitivity to output formats, misleading log-probabilities, and the inability to significantly outperform weak human baselines.





%% file: Sections/limitations.tex
\label{sec:limitations}
Our benchmark reveals key design choices in LM distributional alignment; however, we acknowledge and discuss three limitations of this approach.

\textbf{Scope of surveys topics.} 
Our benchmark and dataset rely on distributions from subjective opinion surveys to capture distributional alignment; however, opinions continuously evolve, surveys may not fully capture diversity and complexity of thought or represent all individuals in that group \citep{durmus2024measuring}, and survey answers may be sensitive to question specificity \citep{berinsky2017} and social desirability bias \citep{yan2021}. While this is an open problem, surveys remain an effective tool in social science for gauging public opinion.

\textbf{Scope of multiple-choice format.}
Our analyses are restricted to opinions expressed in multiple-choice format, which can collapse the nuances of opinions and alter the opinion expressed, as LLMs have also been shown to express different opinions when prompted to respond with open-ended text \citep{wang2024myanswercfirsttoken, lyu2024probabilitiesunveilingmisalignmentevaluating}. 
While eliciting opinions via long-form responses may offer greater ecological validity, we have found complex challenges in studying long-form opinions, such as (1) strict refusal policies that limit an LLM’s ability to generate long-form responses to potentially harmful or generally controversial questions \citep{ouyang2022traininglanguagemodelsfollow,arditi2024refusallanguagemodelsmediated}, (2) challenges in defining the input distribution (e.g., how do users naturally elicit long-form opinions from LMs?) which lead to issues with construct validity, and (3) long-form measurement of opinions encounters the same challenges as the automated evaluation of open-ended text generation, including cost, construct validity, and bias \citep{koo2024benchmarkingcognitivebiaseslarge}. This prevents us from properly benchmarking and making direct comparisons between models. 
Instead, we focus on a high-precision setting of closed-ended survey questions which has several advantages: (1) leveraging established datasets and prior work in this field (e.g., OpinionQA) (2) enabling a more precise, scalable, and reproducible evaluation of LLM performance (3) allowing us to apply existing model calibration techniques.

\textbf{Scope of groups and annotators demographics. }Beyond evaluating six demographic groups for OpinionQA and four demographic groups for NYTBooks, there are many other demographic groups that we have not yet explored. 
Furthermore, we describe the demographics of our human annotators as described in Sec.~\ref{sec:benchmark-human}, which have been limited to the demographic groups we study to enable in-group vs out-group analysis. This slightly limits the representation range in the demographics of our human annotators.

\textbf{Harms of model steerability.}
Studying model steerability towards specific demographic groups can have extreme negative downstream effects, particularly if used to systematically generate misinformation, persuade users to adopt certain opinions, or perpetuate harmful stereotypes. Thus, it is important to acknowledge the risks of model steerability and ensure that model-generated responses are closely monitored in real-world deployments or field studies.

Finally, we do not provide a metric, numerical threshold, or provable statistical test that determines when a system is safe to deploy. It is clear this distributional alignment task is highly context-dependent and socially nuanced, suggesting a one-size-fits-all metric may be more harmful than helpful. 

%% file: Sections/ethical_considerations.tex
\label{sec:ethical_considerations}

While our benchmark (Tab. \ref{tab:main-a}) optimizes for steerability, we caution against blindly optimizing for this metric without considering the harms and limitations of doing so. We advise practitioners to identify when their models misrepresent specific groups and uphold stereotypes as we did in Sec. \ref{sec:exp-implications}, either by collecting disaggregated evaluation metrics to explicitly account for potential discrepancies between groups \citep{barocas2021designingdisaggregatedevaluationsai}, or other metrics that measure LLM simulations' susceptibility to caricature \citep{cheng2023compostcharacterizingevaluatingcaricature,liu2024evaluating}.

A potential risk of our benchmark is that by simulating the distributional opinions of demographic groups, we may inadvertently encourage the use of LLMs to simulate humans. Thus, we emphasize that our benchmark is used only as a discovery mechanism to quantify model capabilities and limitations in distributional alignment. Our objective is to facilitate a deeper understanding of the capabilities and limits of LLMs in emulating human behavior and to ultimately determine if and how we develop such technologies.

%% file: Sections/appendix_acl.tex
\section{Appendix}

\subsection{Significance of Distributional Alignment}
LLM outputs that align with human preferences are becoming increasingly popular (e.g., RLHF). Typically, when researchers evaluate which group a model best aligns with, they compare average responses. However, this only measures alignment to the preference of the majority group. In many subjective tasks, it's crucial for a model to represent the entire spectrum of opinions.

For example, consider the following survey question in the PEW (question ID: $\texttt{GODMORALIMP\_W92}$).

\begin{questionbox}
Regardless of your own religious beliefs, how important, if at all, do you think it is for a person to believe in God in order to be considered good and moral?

\begin{enumerate}[label=\Alph*.]
  \item Important.
  \item Not Important.
\end{enumerate}

\end{questionbox}
The PEW survey respondents responded with $\{A: 54.8\%, B: 45.2\%\}$. If we query the language model and receive a response of ``A. Important'' (the highest likelihood answer choice), the model may appear to be highly aligned if our evaluation only considers the most likely human response. However, this ignores the 45\% of people responding with the minority choice, ``B. Not Important.'' A clearer assessment of this task would be if the language model can represent the 54.8\%, and 45.2\% split, as this would show the model understands the underlying heterogeneity of views on this topic. Evaluating in this way provides a clearer and more precise measure of the model's alignment with human preferences.

\subsection{Temperature Scaling}
Temperature scaling is a post-processing technique to make neural networks calibrated \citep{guo2017calibrationmodernneuralnetworks}. To implement temperature scaling for our work, we find the minimum temperature value that results in the smallest total variation between the reference probabilities and model probabilities. In this setting, the temperature-scaled scaled log-probabilities have access to the ground truth opinion distribution and aim to find $\tau$ that minimizes the total variation. We calculate a new value of $\tau$ per dataset and steering method using the following formula. 
\begin{equation}
\min_{\tau} \frac{1}{2} \left\lVert y_{g, q} - \hat{y}_{g, q, \text{norm}}^{\frac{1}{\tau}} \right\rVert_1 .
\end{equation}

\begin{figure*}
  \centering
  \includegraphics[width=\linewidth]{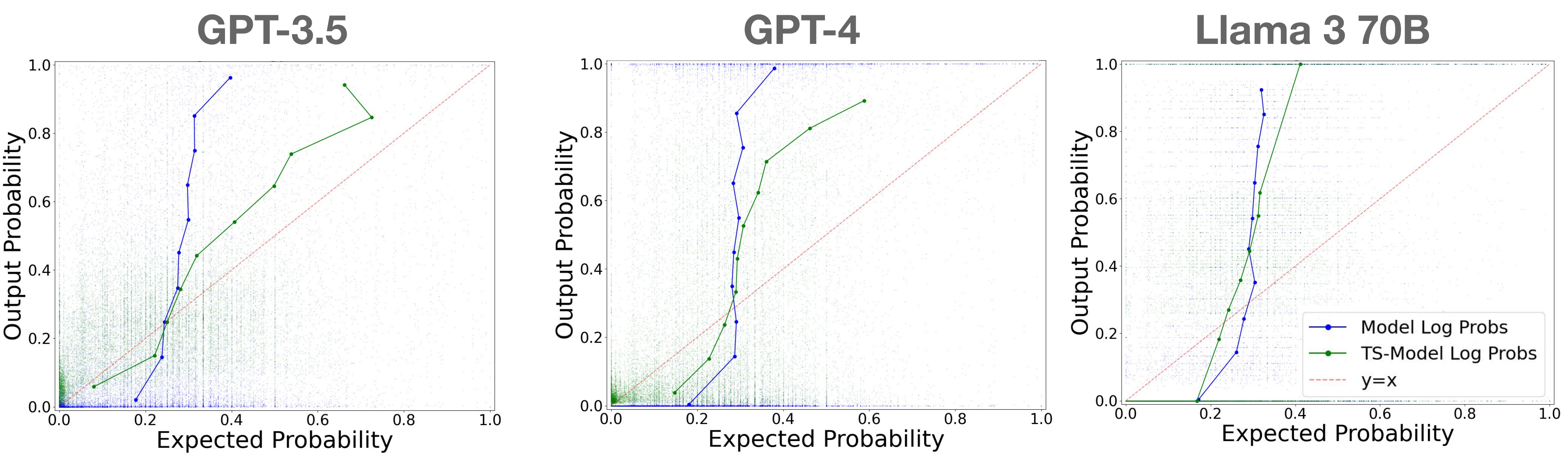}
  \caption{\textbf{Temperature Scaling calibration curves for the survey setting}. We find that temperature scaling improves results for GPT models, but not for Llama3-70b. Results are averaged across the OpinionQA and NYTimes Datasets.}
  \label{fig:temperature_scaling_survey}

\end{figure*}

Applying temperature scaling to the biased coin flip model log-probabilities improves performance. When extended to opinion surveys, we see much fewer improvements in the temperature-scaled model log-probabilities for Llama-3-70B, but improvements for GPT-3.5 and GPT-4. We report the distributional alignment performance in Tab.~\ref{tab:main-a}, plot calibration curves in Fig.~\ref{fig:temperature_scaling_survey}, and report ECE in Tab.~\ref{tab:temp_scale-ECE}).

\begin{table}[h!]
\centering
\begin{tabular}{|l|c|c|}
\hline
\textbf{Model}    & \textbf{TS-Log-p} & \textbf{Log-p} \\ \hline
Llama3-70B        & 0.11                    & 0.13                 \\ \hline
GPT-4             & 0.07                    & 0.28                 \\ \hline
GPT-3.5           & 0.06                    & 0.20                 \\ \hline
\end{tabular}
\caption{ECE (Expected Calibration Error) values for Llama 3 70B, GPT-4, and GPT-3.5.}
\label{tab:temp_scale-ECE}  
\end{table}

\subsection{GlobalOpinionQA}

In this section, we describe additional experiments on a third dataset, GlobalOpinionQA \citep{durmus2024measuring} to reinforce the validity of our findings.

GlobalOpinionQA consists of questions and answers from two cross-national surveys, World Values Survey and PEW Global Attitudes Survey. It is aimed at capturing diverse perspectives on global issues across various countries and is inspired by ~\citet{santurkar2023opinions}. For example, consider this survey question that demonstrates how countries can highly differ in their distributional responses to these survey questions.
\begin{questionbox}
Do you personally believe that drinking alcohol is morally acceptable, morally unacceptable, or is it not a moral issue?

\begin{enumerate}[label=\Alph*.]\item Morally acceptable
\item Morally unacceptable
\item Not a moral issue
\item Depends on the situation 
\end{enumerate}

\end{questionbox}
\begin{answerbox-pakistan}
\begin{enumerate}[label=\Alph*.]
\item Morally acceptable: 0.01
\item Morally unacceptable: 0.95
\item Not a moral issue: 0.02
\item Depends on the situation: 0.02
\end{enumerate}

\end{answerbox-pakistan}

\begin{answerbox-japan}
\begin{enumerate}[label=\Alph*.]
\item Morally acceptable: 0.67
\item Morally unacceptable: 0.06
\item Not a moral issue: 0.25
\item Depends on the situation: 0.02
\end{enumerate}

\end{answerbox-japan}

\begin{answerbox-britain}
\begin{enumerate}[label=\Alph*.]
\item Morally acceptable: 0.39
\item Morally unacceptable: 0.09
\item Not a moral issue: 0.47
\item Depends on the situation: 0.05
\end{enumerate}

\end{answerbox-britain}

\textbf{Pre-processing.} We filter this dataset for the top 100 questions with the highest disagreement between pairs of countries as measured by the distance between the text embedding \citep{gao2021simcse} of the questions. We only consider survey responses from countries that have at least 600 other responses to ensure a large enough pool of questions to pull few shot examples from, reducing the total number of countries in the dataset from 138 to 19.

\textbf{Results.} Our new results are consistent with those in our paper: (1) verbalize remains the optimal distribution estimation method (Tab.~\ref{tab:main-globalvalues}) and (2) there exists a knowledge-to-simulation gap between verbalize and sequence in all 5 models (Tab.~\ref{tab:globalvalues_simgap}). We also separate the results of persona steering and few shot steering and report them in Tab.~\ref{tab:main-globalvalues-persona} and Tab.~\ref{tab:main-globalvalues-fewshot}, respectively. When comparing the persona steered and few shot steered results, we see considerable improvement with few shot steering, corroborating our implications regarding the implications of persona steering and using few shot examples when possible (Sec.~\ref{sec:exp-implications}, Sec.~\ref{sec:discussion}).

\begin{table}[ht!]
  \centering
  \begin{minipage}{\linewidth}
    \centering
    \begin{tabular}{ll}
      \toprule
      Model     & $\mathcal{A}(Y,\hat{Y}_{\mathcal{S}, \mathcal{O}})$  \\
      \midrule
\rowcolor{lightgray} Anthropic Opus (V) & 0.241 $\pm$ 0.015 \\
\rowcolor{lightgray} GPT-4 (V) & 0.279 $\pm$ 0.016 \\
\rowcolor{lightgray} Llama 3 70B (V) & 0.281 $\pm$ 0.016 \\
\rowcolor{lightgray} Anthropic Haiku (V) & 0.293 $\pm$ 0.016 \\
Anthropic Opus (Seq) & 0.301 $\pm$ 0.020 \\
GPT-4 (TS-Log-p) & 0.303 $\pm$ 0.016 \\
Llama 3 70B (Seq) & 0.346 $\pm$ 0.020 \\
Anthropic Haiku (Seq) & 0.351 $\pm$ 0.016 \\
\rowcolor{lightgray} GPT-3.5 (V) & 0.356 $\pm$ 0.019 \\
GPT-4 (Seq) & 0.359 $\pm$ 0.021 \\
GPT-3.5 (TS-Log-p) & 0.381 $\pm$ 0.015 \\
GPT-3.5 (Seq) & 0.399 $\pm$ 0.018 \\
GPT-3.5 (Log-p) & 0.442 $\pm$ 0.022 \\
Llama 3 70B (Log-p) & 0.455 $\pm$ 0.021 \\
GPT-4 (Log-p) & 0.484 $\pm$ 0.023 \\
Llama 3 70B (TS-Log-p) & 0.491 $\pm$ 0.024 \\
            \midrule

Discretization Error (Seq) & 0.092 $\pm$ 0.004 \\
            Uniform & 0.486 $\pm$ 0.019 \\
            Majority Vote & 0.692 $\pm$ 0.035 \\
            Minority Vote & 0.854 $\pm$ 0.027 \\
      \bottomrule
    \end{tabular}
    \caption{\textbf{Distributional Alignment Task on GlobalOpinionQA}. Models ranked based on mean total variation. Models highlighted in gray and with (V) have a distribution expression method of directly verbalizing the distribution in a JSON format ($\mathcal{O} = \texttt{Verbalize}$). Models not highlighted represent samplers, where (Seq) represents the 30-token sequential distribution output ($\mathcal{O} = \texttt{Sequence}$), (Log-p) represents $\mathcal{O} = \texttt{Model Log-probabilities}$, and (TS-Log-p) represents $\mathcal{O} = \texttt{Temperature Scaled Model Log-probabilities}$. }
    \label{tab:main-globalvalues}  
    
  \end{minipage}%
\end{table}

\begin{table}[h!]
  \centering
  \begin{minipage}{\linewidth}
    \centering
    
\begin{tabular}{lc}
      \toprule
      Model     & Percent Error (\%)     \\
      \midrule
      GPT-3.5 Turbo & 12.15\%  \\
      Claude: Haiku & 19.86\%   \\
      Llama 3 70B Instruct & 22.95\%    \\
      Claude: Opus & 24.88\% \\
        GPT-4 & 28.68\%  \\
      \bottomrule
\end{tabular}
\caption{\textbf{Knowledge-to-Simulation Gap of GlobalOpinionQA(Eq.~\ref{eq:knowledge_to_sim})}. The simulation penalty measures the percent error increase in total variation between the 30-token sequential distribution output and the verbalization of knowledge.}
\label{tab:globalvalues_simgap}

  \end{minipage}%
\end{table}

\begin{table}[ht!]
  \centering
  \begin{minipage}{\linewidth}
    \centering
    \begin{tabular}{ll}
      \toprule
      Model     & $\mathcal{A}(Y,\hat{Y}_{\mathcal{S}, \mathcal{O}})$  \\
      \midrule
      \rowcolor{lightgray} Anthropic Opus (V) & 0.275 $\pm$ 0.021 \\
\rowcolor{lightgray} GPT-4 (V) & 0.319 $\pm$ 0.022 \\
GPT-4 (TS-Log-p) & 0.325 $\pm$ 0.022 \\
\rowcolor{lightgray} Llama 3 70B (V) & 0.327 $\pm$ 0.024 \\
\rowcolor{lightgray} Anthropic Haiku (V) & 0.333 $\pm$ 0.022 \\
Anthropic Opus (Seq) & 0.338 $\pm$ 0.028 \\
Anthropic Haiku (Seq) & 0.374 $\pm$ 0.021 \\
Llama 3 70B (Seq) & 0.377 $\pm$ 0.028 \\
GPT-4 (Seq) & 0.378 $\pm$ 0.029 \\
\rowcolor{lightgray} GPT-3.5 (V) & 0.397 $\pm$ 0.024 \\
GPT-3.5 (TS-Log-p) & 0.428 $\pm$ 0.018 \\
GPT-3.5 (Seq) & 0.432 $\pm$ 0.020 \\
GPT-3.5 (Log-p) & 0.446 $\pm$ 0.027 \\
Llama 3 70B (Log-p) & 0.462 $\pm$ 0.028 \\
GPT-4 (Log-p) & 0.508 $\pm$ 0.033 \\
Llama 3 70B (TS-Log-p) & 0.546 $\pm$ 0.036 \\
            \midrule

Discretization Error (Seq) & 0.092 $\pm$ 0.004 \\
            Uniform & 0.486 $\pm$ 0.019 \\
      \bottomrule
    \end{tabular}
    \caption{\textbf{Distributional Alignment Task on GlobalOpinionQA: Persona Steering}. Models ranked based on mean total variation. Models highlighted in gray and with (V) have a distribution expression method of directly verbalizing the distribution in a JSON format ($\mathcal{O} = \texttt{Verbalize}$). Models not highlighted represent samplers, where (Seq) represents the 30-token sequential distribution output ($\mathcal{O} = \texttt{Sequence}$), (Log-p) represents $\mathcal{O} = \texttt{Model Log-probabilities}$, and (TS-Logp) represents $\mathcal{O} = \texttt{Temperature Scaled Model Log-probabilities}$. }
    \label{tab:main-globalvalues-persona}  
    
  \end{minipage}%
\end{table}

\begin{table}[ht!]
  \centering
  \begin{minipage}{\linewidth}
    \centering
    \begin{tabular}{ll}
      \toprule
      Model     & $\mathcal{A}(Y,\hat{Y}_{\mathcal{S}, \mathcal{O}})$  \\
            \midrule
      \rowcolor{lightgray}
Anthropic Opus (V) & 0.208 $\pm$ 0.020 \\
      \rowcolor{lightgray}
Llama 3 70B (V) & 0.234 $\pm$ 0.022 \\
      \rowcolor{lightgray}
GPT-4 (V) & 0.237 $\pm$ 0.023 \\
      \rowcolor{lightgray}
Anthropic Haiku (V) & 0.252 $\pm$ 0.023 \\
Anthropic Opus (Seq) & 0.265 $\pm$ 0.025 \\
GPT-4 (TS-Log-p) & 0.280 $\pm$ 0.024 \\
GPT-4 (Seq) & 0.341 $\pm$ 0.032 \\
Llama 3 70B (Seq) & 0.312 $\pm$ 0.026 \\
\rowcolor{lightgray}
GPT-3.5 (V) & 0.314 $\pm$ 0.026 \\
Anthropic Haiku (Seq) & 0.327 $\pm$ 0.024 \\
GPT-3.5 (TS-Log-p) & 0.334 $\pm$ 0.023 \\
GPT-3.5 (Seq) & 0.366 $\pm$ 0.027 \\
GPT-3.5 (Log-p) & 0.440 $\pm$ 0.032 \\
Llama 3 70B (Log-p) & 0.449 $\pm$ 0.031 \\
GPT-4 (Log-p) & 0.461 $\pm$ 0.033 \\
Llama 3 70B (TS-Log-p) & 0.437 $\pm$ 0.031 \\
            \midrule
            
Discretization Error (Seq) & 0.092 $\pm$ 0.004 \\
            Uniform & 0.486 $\pm$ 0.019 \\
      \bottomrule
    \end{tabular}
    \caption{\textbf{Distributional Alignment Task on GlobalOpinionQA: Few Shot Steering}. Models ranked based on mean total variation. Models highlighted in gray and with (V) have a distribution expression method of directly verbalizing the distribution in a JSON format ($\mathcal{O} = \texttt{Verbalize}$). Models not highlighted represent samplers, where (Seq) represents the 30-token sequential distribution output ($\mathcal{O} = \texttt{Sequence}$), (Log-p) represents $\mathcal{O} = \texttt{Model Log-probabilities}$, and (TS-Logp) represents $\mathcal{O} = \texttt{Temperature Scaled Model Log-probabilities}$. }
    \label{tab:main-globalvalues-fewshot}  
    
  \end{minipage}%
\end{table}

\subsection{NYT Books Dataset Construction}
\begin{figure}[ht!]
  \centering
  \includegraphics[width=\linewidth]{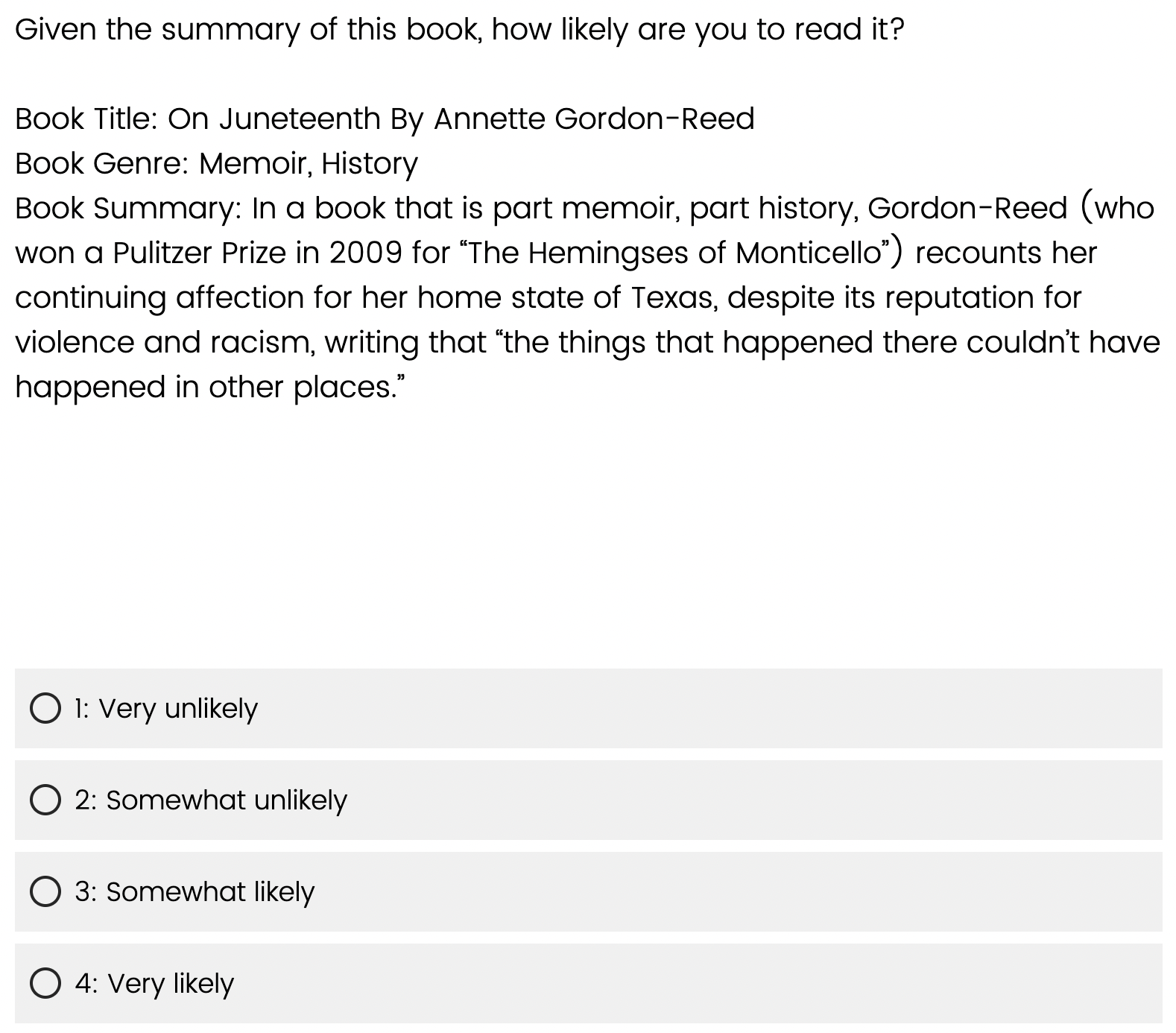}
  \caption{\textbf{NYT Books Annotation Example Task.}}
  \label{fig:nyt_example}
\end{figure}

\begin{figure}[ht!]
  \centering
  \includegraphics[width=\linewidth]{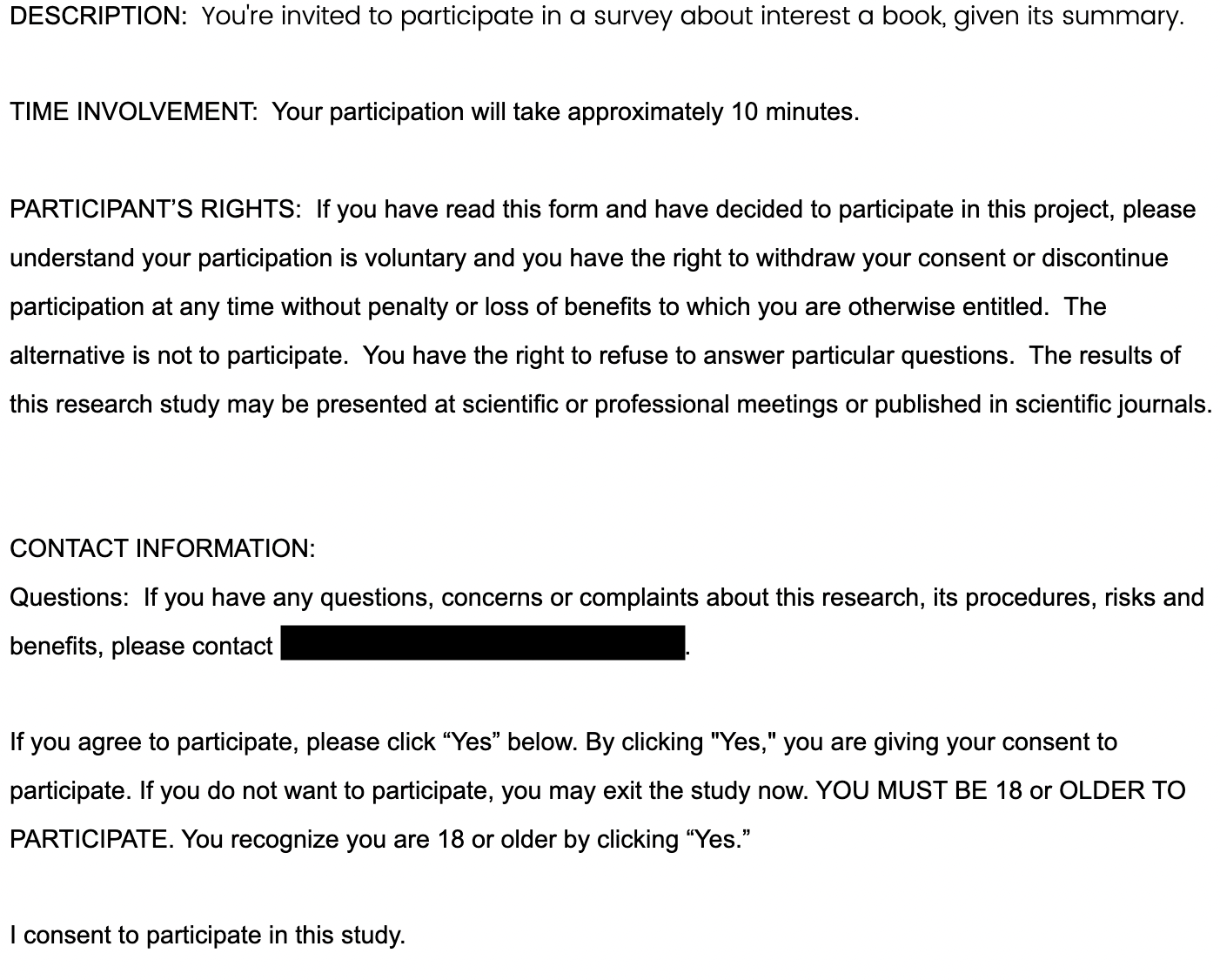}
  \caption{\textbf{NYT Books Annotator Consent Form.}}
  \label{fig:nyt_consent}
\end{figure}

\textbf{Annotators.}
From this data collection process, we surveyed 131 Male, 206 Female, 165 Democrat, and 172 Republican annotators, resulting in 18 annotations per book per demographic group. 
In Fig \ref{fig:nyt_example}, we show an example annotation question. In this example, annotators are given a book title and its
corresponding author, book summary, and genre. Then they provide a 4-point Likert rating to the question, ``Given the summary of this book, how likely are you to read it?"  
From these annotations, we calculated an opinion distribution over the 4 Likert ratings for each book. All crowd workers are sourced on Prolific, filtered for English fluency, selected from a pool of annotators who pass an attention check 93\% of the time and are paid \$12 per hour (this amount is well over the federal minimum wage of \$7.25). The consent form shown to annotators is shown in Fig
\ref{fig:nyt_consent}. 

It should be noted that articles from the New York Times are in English and protected under copyright, but this research is performed in the public interest under GDPR and the excerpts are collected under fair use exemption. When releasing our dataset, we do not release any of the texts of the article, only links to the New York Times website to respect copyrights. Furthermore, the annotation data we collect is not personally identifiable as it consists only of opinion ratings of books. 

In addition to the disagreement analysis we conduct in Sec. \ref{sec:benchmark-datasets}, we calculate the Cohen's kappa of opinions \textit{between} demographic groups. We find that Cohen's kappa between Democrats and Republicans is $0.05$, indicating little to no agreement. The Cohen's kappa between Men and Women is $0.15$, indicating a small amount of agreement. 

\subsection{OpinionQA} We used the OpinionQA \citet{santurkar2023opinions} accessed via the CC-BY 4.0 license, written in English, as they survey US participants.  In their steerability analysis, they create a smaller set of 500 contentious questions where the subgroups frequently disagree. We follow suit and randomly sample 100 questions from this set to obtain questions spanning topics such as science, politics, and personal relationships.

\subsection{Human Baseline Annotations}

Using the Prolific platform, we crowdsource human annotations for the distributional alignment task. 
We restrict the demographics of annotators to match the groups that we study as it provided us an opportunity to do an in-group vs out-group analysis. We restrict our annotators to Democrat (73\%), Republican (27\%), Male (33\%), Female (67\%), White (84\%), and Black (16\%) for the OQA annotations. For the NYT annotations, our annotators are Democrat (69\%), Republican (31\%), Male (37\%), and Female (63\%). Each question from the OQA and NYT datasets is annotated four times. We compensate the workers at a rate of \$12 per hour  (this amount is well over the federal minimum wage of \$7.25). The consent form shown to annotators is similar to that shown for the NYT Books annotation collection (Fig.
\ref{fig:nyt_consent})---the only difference is the ``description''. 

\begin{figure}[ht!]
  \centering
  \includegraphics[width=\linewidth]{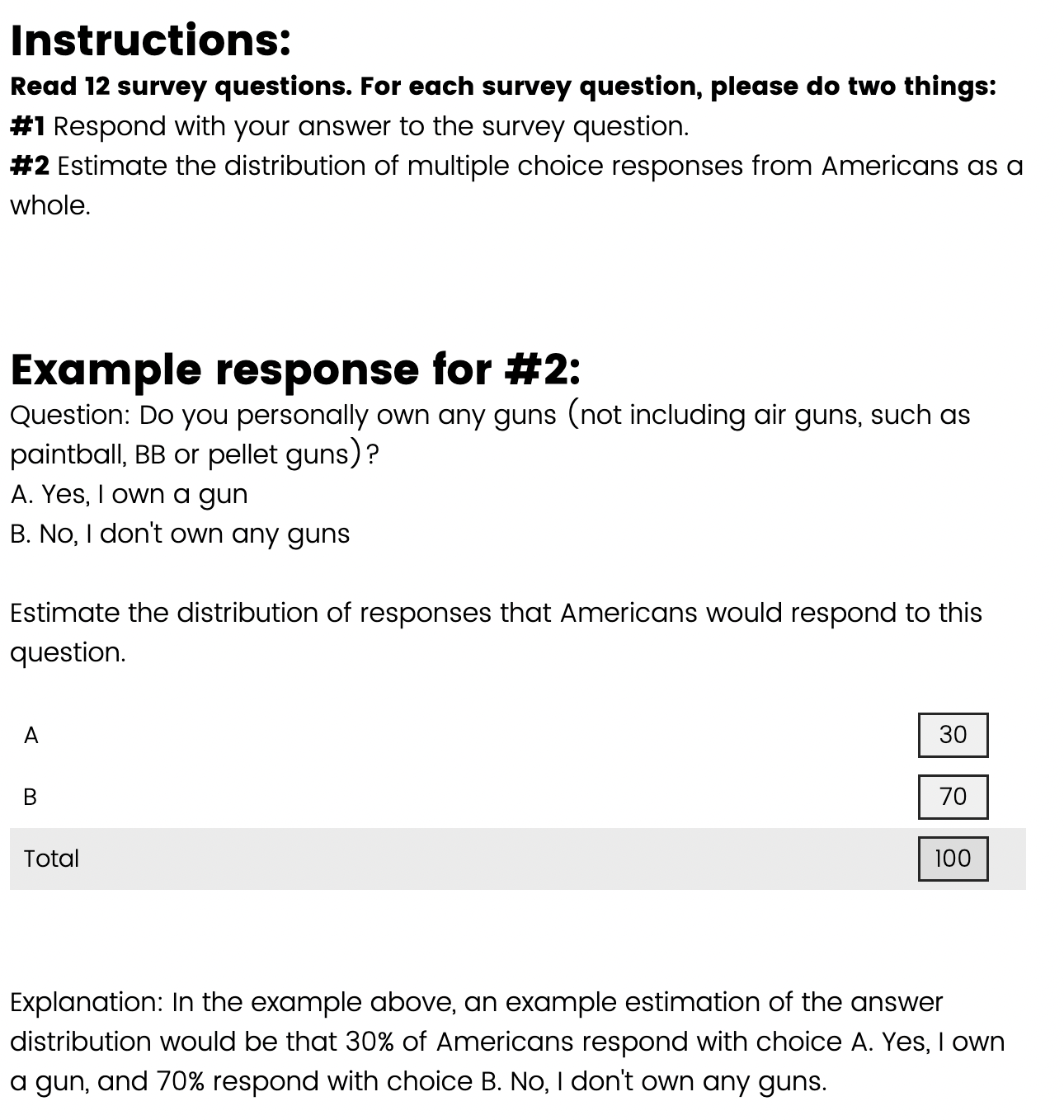}
  \caption{\textbf{Distributional Alignment Instructions and Example Task: No Steering}.}
  \label{fig:DA_nosteer_example}
\end{figure}
\begin{figure}[ht!]
  \centering
  \includegraphics[width=\linewidth]{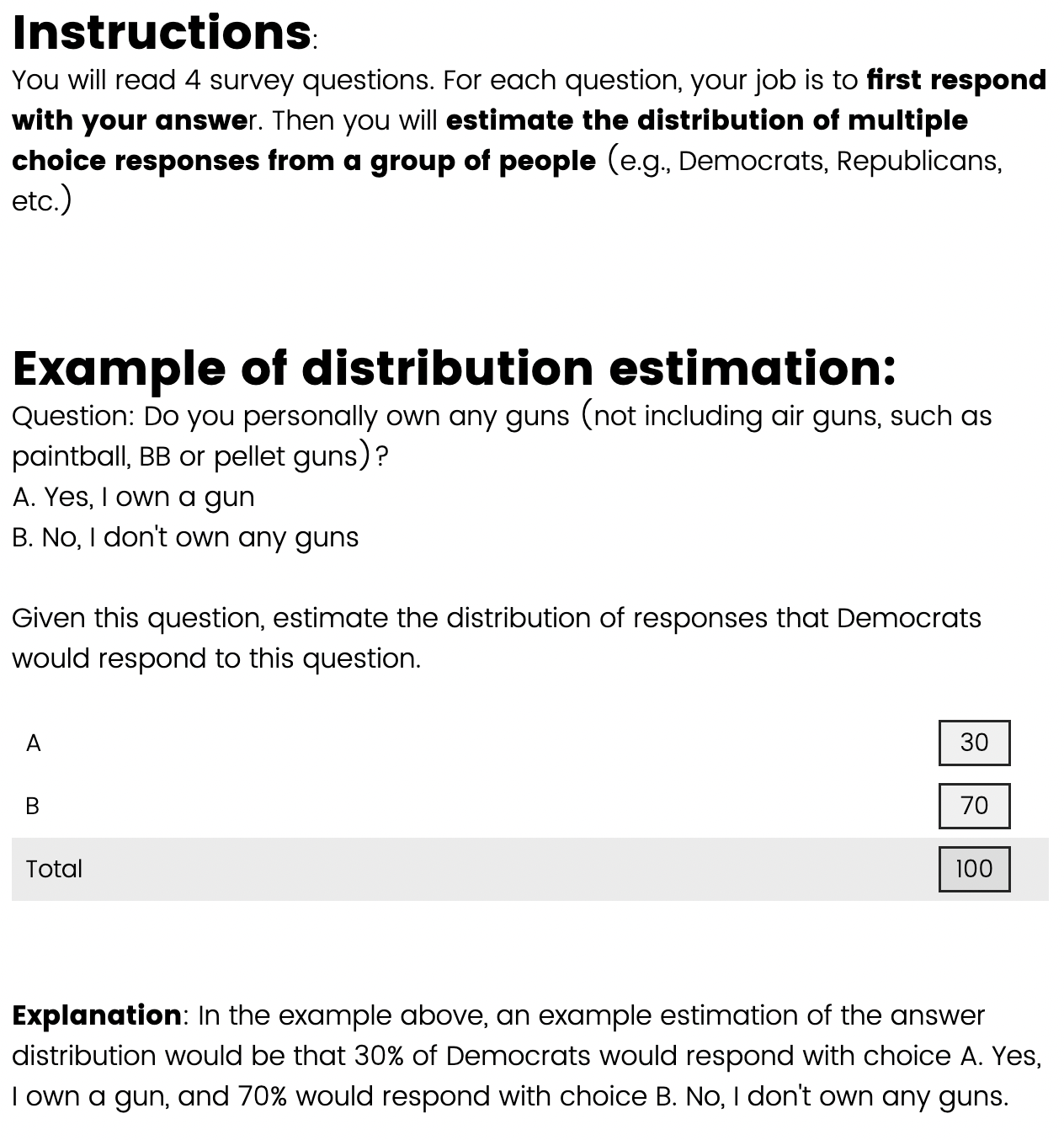}
  \caption{\textbf{Distributional Alignment Instructions and Example Task: Persona Steering}.}
  \label{fig:DA_persona_Example}
\end{figure}
\begin{figure}[ht!]
  \centering
  \includegraphics[width=\linewidth]{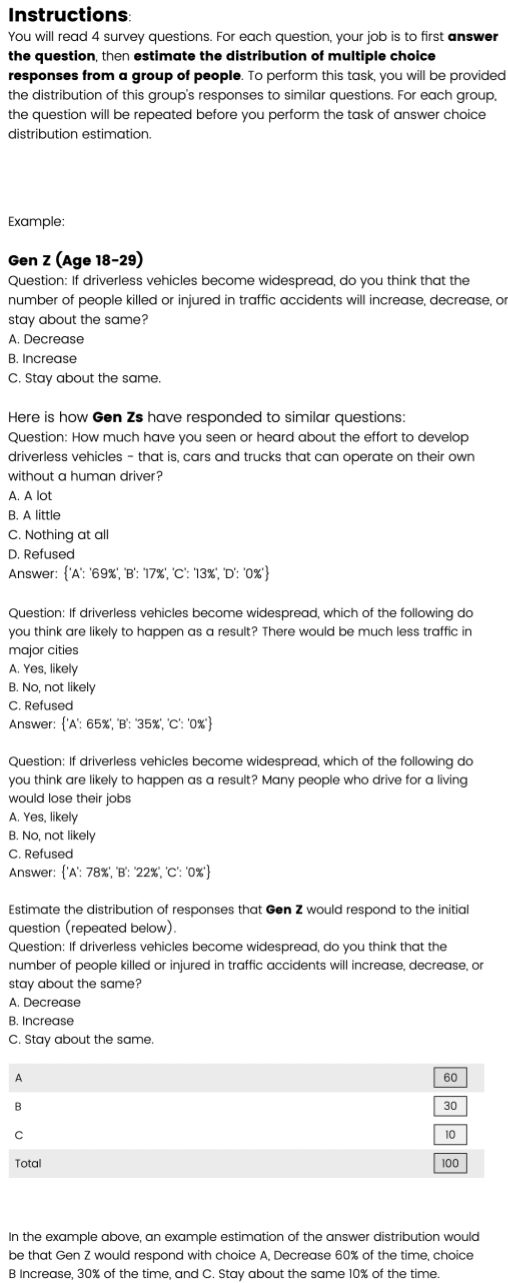}
  \caption{\textbf{Distributional Alignment Instructions and Example Task: Few Shot Steering}.}
  \label{fig:DA_FewShot_Example}
\end{figure}

Next, we show the instructions given to participants and a demo of how the distributional alignment task is completed.  In Fig. \ref{fig:DA_nosteer_example}, we show the instructions for no steering. In this example, annotators are instructed to estimate the distribution of multiple choice responses from Americans as a whole.  
In Fig. \ref{fig:DA_persona_Example}, we show the instructions for persona steering. In this specific example, annotators are instructed to simulate a Democrat, but this group can be any demographic group of interest. 
Finally, in Fig. \ref{fig:DA_FewShot_Example}, we show the instructions and an example annotation for few shot steering. In this specific example, humans are tasked with simulating the views of Gen Z and given examples of how this group has responded to similar questions on driverless vehicles (participants are given 5 examples, but only 3 are shown in this figure). 

\textbf{Quality control.}
To ensure annotation quality, we limit the task to workers with English fluency and ask survey participants to answer a reading attention check question which our annotators achieved 93\% on.

Prior work has found that in-group representations and out-group imitations from human participants
are different and that there exists a misperception of partisan polarization~\citep{wang2024large,levendusky}. 
Thus, we compared in-group and out-group representations in opinion distribution estimation. In this context, ``in-group'' refers to Democrats simulating Democrats, and ``out-group'' refers to Democrats simulating Republicans. ``In-group'' can also refer to a group of men simulating other men and ``out-group'' refers to a group of men simulating a group of women. 

While there is more difference on average over all subgroups in persona steering between the in-group and out-group, these differences are not statistically significant (Tab.~\ref{tab:in_out_group}). However, at the sub-group level, we do see statistically significant evidence of a \textit{false polarization phenomenon}, where there exists a misperception of in-group and out-group representations for Democrats and Republicans, but this difference is not statistically significant for other demographic groups (Men-Women, Black-White). It should be noted that this false polarization phenomenon only exists with persona steering (i.e., when prompted to behave as a certain demographic group), not when few shot examples for that demographic group are provided. All results are reported in Fig.~\ref{fig:in_out_subgroups}.

\begin{figure}[ht!]
  \centering
  \includegraphics[width=\linewidth]{Figures/ingroup_outgroup (3).pdf}
  \caption{\textbf{In group vs out group alignment of human annotators}. On the x-axis, we plot the total variation between the human annotations and the ground truth opinion distribution, separated by in-group and out-group. A triangle denotes in-group alignment, meaning a Democrat is emulating another Democrat. A circle denotes out-group alignment, where a Democrat is emulating a Republican. In blue, we highlight the instances of false polarization phenomenon, where Democrats are much more aligned when emulating the in-group than the out-group.}
  \label{fig:in_out_subgroups}
\end{figure}

\begin{table}[ht]
  \centering
  \begin{tabular}{llll}
    \cmidrule{1-4}
    Dataset & Steering      & In/Out & TV  \\
    \midrule
    OQA & Persona & In    &   0.298 $\pm$ 0.013  \\

    OQA & Persona &  Out   &  0.321 $\pm$ 0.014  \\
\midrule
    NYT & Persona & In    &   0.281 $\pm$ 0.010  \\
    NYT & Persona &  Out   &  0.273 $\pm$ 0.010  \\
    \midrule
    OQA & Few Shot & In    &  0.283 $\pm$ 0.013  \\
    OQA & Few Shot & Out    &  0.278 $\pm$ 0.013 \\
\midrule
    NYT & Few Shot & In    &  0.244 $\pm$ 0.008  \\
    NYT & Few Shot & Out    &  0.246 $\pm$ 0.009 \\    

    \bottomrule
  \end{tabular}
  \caption{In-group vs. Out-group performance in human annotators, averaged over all demographic groups. }
  \label{tab:in_out_group}
\end{table}

\subsection{Model Log Probabilties of each Token in the Sequence of Biased Coin Flips}

\begin{figure}[ht!]
  \centering
  \includegraphics[width=\linewidth]{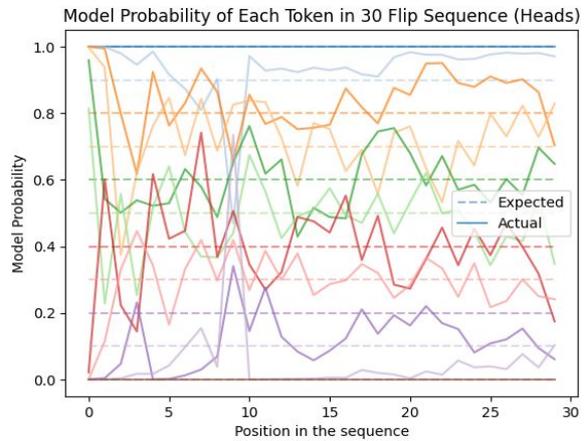}
  \caption{\textbf{Model Probability of Each Token in 30 Flip Sequence} (Heads).}
  \label{fig:BiasedCoinFlip_Seq_ModelLogProbs_H}
\end{figure}

In this section, we provide additional analysis on the Biased Coin Flip experiment, specifically regarding the sequence of biased coin flips. We tested the OpenAI \texttt{gpt-4} model, access on March 2024, which points to \textt{gpt-4-0613}, a snapshot of \texttt{gpt-4} from June 13th 2023. 
In Fig.~\ref{fig:BiasedCoinFlip_Seq_ModelLogProbs_H}, we plot the model log probabilities of each token in the sequence to better understand the conditional probabilities.

\subsection{Few Shot Steering}
\label{sec:appendix-fewshot}
To perform few shot steering, for each question, we calculate the top 10 most similar questions or books as identified from their text embeddings 
\citep{gao2021simcse}. We found that some of these questions may be near-duplicates, especially in opinionQA where a survey may include many variants of the same question. Thus, we filter for the hardest similar examples, namely the top 5 questions most distinct in output distribution. 
This ensures we provide topically coherent and similar examples while avoiding cases where the model simply copies the distribution from the few shot examples. We pass these 5 examples in as contextual distributional information to aid in the distribution estimation.

\subsection{Additional Model Information}

In this section, we discuss each model and how it was accessed. All GPT models and their log-probabilities were accessed using the OpenAI API, namely $\texttt{GPT-3.5-Turbo-0125}$ and $\texttt{gpt-4-0613}$. 
Both Anthropic Haiku and Anthropic Opus were accessed using the Anthropic API which does not provide log-probabilities to users. Finally, Meta's $\texttt{Meta-Llama-3-70B-Instruct}$ was accessed via Huggingface. Since most models were accessed via API, the GPU hours accrewed come from running inference on Meta's Llama-3-70B-Instruct---this amounted to less than 1 hour. 

While we tested additional models listed in Tab.~\ref{tab:model_performance}, we found that they struggled to follow the distribution estimation method prompt, particularly for that of sequence and verbalize.

\begin{table}[ht!]

  \label{sample-table}
  \centering
  \begin{tabular}{lll}
  
    \cmidrule{1-3}
    Model & Seq & Verbalize \\
    \cmidrule{1-3}
    Llama-3-8B & 45\% & 40\% \\
    Llama-2-70B & 3\% & 97\% \\
    Llama-2-13B & 11\% & 83\% \\
    Llama-2-7B & 10\% & 52\% \\
    Deepseek-coder-1.3B & 0\% & 2\% \\
    Deepseek-coder-6.7B & 0\% & 0\% \\
\bottomrule
\end{tabular}
\caption{\textbf{Open models struggle to follow the prompt for the distribution expression methods, sequence, and verbalize.} In this table, we report the success rate that additional open models had in following the prompt instructions. There was limited success with these smaller models; thus, we opted for the 5 models included in the main paper. }
\label{tab:model_performance}
\end{table}

\subsection{Persona-steered LLMs \textit{and humans} are susceptible to stereotyping.}
\begin{figure*}[ht!]
  \centering
  \includegraphics[width=\linewidth]{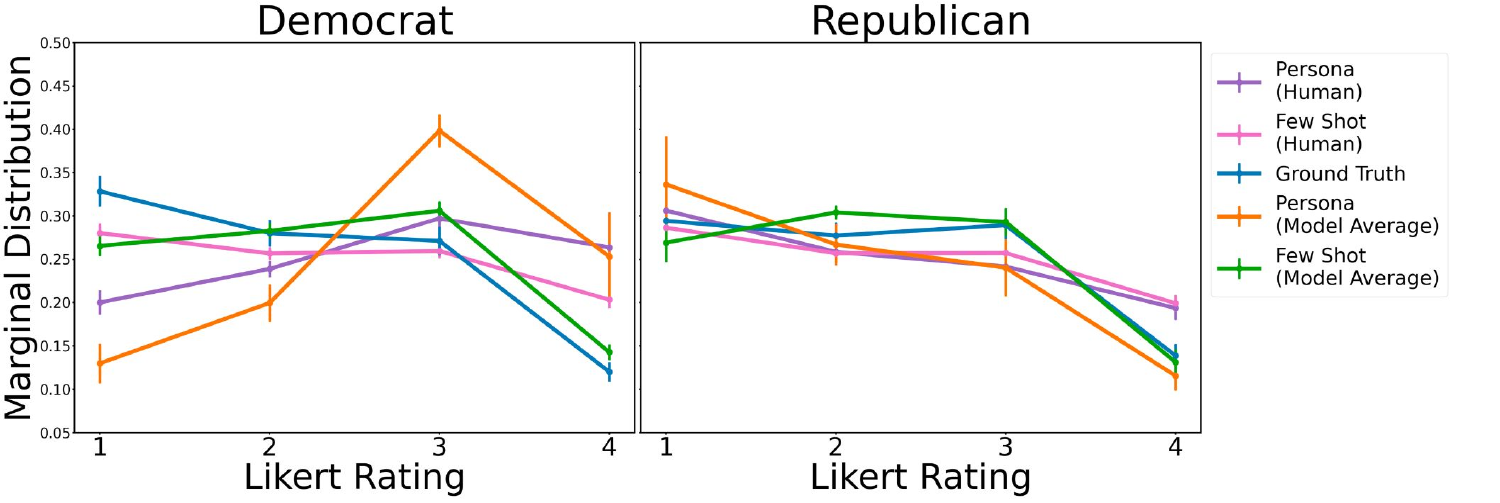}
  \caption{ \textbf{Persona-steered LLMs and humans produce stereotypical results.} We plot the marginal distribution for each Likert Rating (1-4) corresponding to the answer to the following question: ``How likely are you to read this book?'' A Likert rating of 1 refers to ``Very unlikely'' and a Likert rating of 4 refers to ``Very likely''. 
  We report the average marginal distribution over 5 models and humans steered towards Democrats and Republicans with persona steering (orange) and few shot steering (green). In blue, we plot the reference human reference for Democrat and Republican annotators. All of the data is averaged over 235 books in the NYT Books dataset. In purple we plot the persona steered humans, and in pink we plot the few shot steered humans. We observe that persona steered humans see a similar stereotypical effect, although smaller. }
  \label{fig:NYT_stereotyping_humans}
\end{figure*}

We supplement Fig.~\ref{fig:NYT_stereotyping} which depicts the marginal distribution of the answer ratings in the NYT Books dataset with average human ratings (Fig.~\ref{fig:NYT_stereotyping_humans}). While previously, we demonstrated that models prompted with a Democrat persona tend to simulate humans that are more likely to read books, we see that humans have a similar stereotypical effect, although smaller than that of models. 

\subsection{Additional Distribution Estimation Method Related Works} 
In this section, we engage more substantially with prior work relating to the distribution estimation method (Sec.~\ref{sec:benchmark-estim-method}). Prior work has found that language models can either be queried for statement
probabilities (e.g., verbalize or other prompt-based techniques) or probed for internal representations of truthfulness (i.e., model log-probabilities). Past work has found that these two representations can sometimes disagree~\citep{hu2023prompting,mondal2024largelanguagemodelsexhibit,liu-etal-2023-cognitive}. Earlier work by ~\citet{hu2023prompting} finds that direct probability measurements generally field better or similar performance when compared to prompting techniques on tasks such as word prediction, semantic plausibility, and syntax completion in a zero-shot setting. Our work differs in that our tasks involve scenarios with uncertainty, we evaluate RLHF-ed LMs, and we provide few shot examples of the task. This is valuable as LMs with RLHF have been shown to produce conditional probabilities that are poorly calibrated~\citep{kadavath2022languagemodelsmostlyknow}.

Furthermore, the estimated distribution of the model's opinion has been shown to be sensitive to other factors including rewording \citep{Bisbee_Clinton_Dorff_Kenkel_Larson_2024} or negating questions \citep{ceron2024promptbrittlenessevaluatingreliability}, shuffling the answer choice order \citep{dominguezolmedo2024questioningsurveyresponseslarge}, and when converting multiple-choice surveys to free text \citep{röttger2024politicalcompassspinningarrow,lyu2024probabilitiesunveilingmisalignmentevaluating,wang2024myanswercfirsttoken}.

\subsection{Distributional Distance Metric: Total Variation}
Equ.~\ref{eq:distributional_alignment} represents the total variation \emph{distance}, a measure for comparing the distance between two probability distributions (see Prop 4.2 in \citet{LevinPeresWilmer2006}). While KL divergence can be used to compare distributions' differences, we selected total variation distance for its interpretability and relevance to our setting. 
One notable reason why KL divergence is unsuitable for comparing survey responses from people with those from the model is that KL divergence becomes infinite when one distribution has zero probability events while the other does not. This naturally occurs in survey samples when no respondents select a particular answer choice, or in our distribution estimation methods, sequence and verbalize, which can generate zero estimated probabilities for certain answer choices.

Consider two probability distributions $P$ and $Q$ over the discrete space of multiple-choice options $\{A, B, C\}$, where 
\begin{align*}
P(A) &= 0.6, \quad P(B) = 0.35, \quad P(C) = 0.05 \\
Q(A) &= 0.6, \quad Q(B) = 0.40,  \quad Q(C) = 0.
\end{align*}

$$D_{KL}(P || Q) = \sum_{i=1}^{3} P(i) \log \left( \frac{P(i)}{Q(i)} \right)$$

The final term ($i=3$) contains $0.05 \log \left( \frac{0.05}{0} \right)$ which goes to infinity ($\log(0) = -\infty $).

\begin{align*}
\text{TV}(P, Q) &= \frac{1}{2} \sum_{i=1}^{3} ||P(i) - Q(i)||_1 = 0.05.
\end{align*}

Here total variation distance is a small value, reflecting that these distributions are very similar.

\subsection{Full Results of Each Steering Method}
In Tab.~\ref{tab:all_results} we report the results of each steering method (No Steering, Persona Steering, and Few Shot Steering) for OpinionQA and NYTBooks. 
\onecolumn

\begin{longtable}{|c|ll|ll|}
      \hline
    \multirow{18}{*}{\rotatebox{90}{\textbf{No Steering}}} 
    & \multicolumn{2}{c|}{\cellcolor{white}\textbf{OQA}} & \multicolumn{2}{c|}{\cellcolor{white}\textbf{NYT}} \\
    \cline{2-5}
    & \cellcolor{white}Claude: Haiku (Seq) & \cellcolor{white}0.372 $\pm$ 0.014 & \cellcolor{lightgray}Anthropic Opus (V) & \cellcolor{lightgray}0.210 $\pm$ 0.006   \\
    & \cellcolor{white}GPT-3.5 (TS-Log-p) & \cellcolor{white}0.386 $\pm$ 0.012 & \cellcolor{white}GPT-3.5 (Seq) & \cellcolor{white}0.215 $\pm$ 0.006  \\
    & \cellcolor{white}GPT-4 (Seq) & \cellcolor{white}0.390 $\pm$ 0.014 &  \cellcolor{lightgray}GPT-3.5 (V) & \cellcolor{lightgray}0.219 $\pm$ 0.006   \\
    & \cellcolor{lightgray}GPT-3.5 (V) & \cellcolor{lightgray}0.393 $\pm$ 0.014 & \cellcolor{white}GPT-4 (Seq) & \cellcolor{white}0.220 $\pm$ 0.006  \\
    & \cellcolor{lightgray}Llama 3 70B (V) & \cellcolor{lightgray}0.397 $\pm$ 0.013 & \cellcolor{white}Anthropic Haiku (Seq) & \cellcolor{white}0.226 $\pm$ 0.007   \\
    & \cellcolor{white}GPT-3.5 (Seq) & \cellcolor{white}0.398 $\pm$ 0.014 & \cellcolor{lightgray}GPT-4 (V) & \cellcolor{lightgray}0.229 $\pm$ 0.007  \\
    & \cellcolor{lightgray}GPT-4 (V) & \cellcolor{lightgray}0.400 $\pm$ 0.014 & \cellcolor{lightgray}Llama 3 70B (V) & \cellcolor{lightgray}0.237 $\pm$ 0.007 \\
    & \cellcolor{white}GPT-4 (TS-Log-p) & \cellcolor{white}0.411 $\pm$ 0.012 & \cellcolor{white}Llama 3 70B (Seq) & \cellcolor{white}0.257 $\pm$ 0.009   \\
    & \cellcolor{lightgray}Claude: Haiku (V) & \cellcolor{lightgray}0.427 $\pm$ 0.015 & \cellcolor{lightgray}Anthropic Haiku (V) & \cellcolor{lightgray}0.265 $\pm$ 0.007  \\
    & \cellcolor{white}Claude: Opus (Seq) & \cellcolor{white}0.484 $\pm$ 0.019 & \cellcolor{white}Anthropic Opus (Seq) & \cellcolor{white}0.316 $\pm$ 0.011  \\
    & \cellcolor{lightgray}Claude: Opus (V) & \cellcolor{lightgray}0.451 $\pm$ 0.014 & \cellcolor{white}GPT-4 (TS-Log-p) & \cellcolor{white}0.323 $\pm$ 0.008  \\
    & \cellcolor{white}Llama 3 70B (Seq) & \cellcolor{white}0.501 $\pm$ 0.020 &  \cellcolor{white}GPT-3.5 (TS-Log-p) & \cellcolor{white}0.460 $\pm$ 0.008  \\
    & \cellcolor{white}GPT-3.5 (Log-p) & \cellcolor{white}0.540 $\pm$ 0.017 &  \cellcolor{white}Llama 3 70B (TS-Log-p) & \cellcolor{white}0.606 $\pm$ 0.010  \\
    & \cellcolor{white}Llama 3 70B (TS-Log-p) & \cellcolor{white}0.662 $\pm$ 0.019 & \cellcolor{white}Llama 3 70B (Log-p) & \cellcolor{white}0.638 $\pm$ 0.010  \\
    & \cellcolor{white}Llama 3 70B (Log-p) & \cellcolor{white}0.688 $\pm$ 0.018 & \cellcolor{white}GPT-4 (Log-p) & \cellcolor{white}0.689 $\pm$ 0.010  \\
    & \cellcolor{white}GPT-4 (Log-p) & \cellcolor{white}0.714 $\pm$ 0.018 & \cellcolor{white}GPT-3.5 (Log-p) & \cellcolor{white}0.745 $\pm$ 0.009  \\
    \hline

    \multirow{18}{*}{\rotatebox{90}{\textbf{Persona Steering}}}
    & \multicolumn{2}{c|}{\cellcolor{white}\textbf{OQA}} & \multicolumn{2}{c|}{\cellcolor{white}\textbf{NYT}} \\
    \cline{2-5}
    
    & \cellcolor{lightgray}GPT-4 (V) & \cellcolor{lightgray}0.181 $\pm$ 0.007 & \cellcolor{white}GPT-3.5 (Seq) & \cellcolor{white}0.226 $\pm$ 0.007   \\
    & \cellcolor{lightgray}Claude: Haiku (V) & \cellcolor{lightgray}0.222 $\pm$ 0.011 & \cellcolor{lightgray}GPT-3.5 (V) & \cellcolor{lightgray}0.239 $\pm$ 0.007  \\
    & \cellcolor{white}GPT-4 (TS-Log-p) & \cellcolor{white}0.224 $\pm$ 0.012 & \cellcolor{lightgray}GPT-4 (V) & \cellcolor{lightgray}0.248 $\pm$ 0.007  \\
    & \cellcolor{lightgray}Llama 3 70B (V) & \cellcolor{lightgray}0.226 $\pm$ 0.009 & \cellcolor{lightgray}Llama 3 70B (V) & \cellcolor{lightgray}0.257 $\pm$ 0.008   \\
    & \cellcolor{lightgray}Anthropic Opus (V) & \cellcolor{lightgray}0.238 $\pm$ 0.012 &\cellcolor{white}GPT-4 (Seq) & \cellcolor{white}0.264 $\pm$ 0.009  \\
    & \cellcolor{white}GPT-4 (Seq) & \cellcolor{white}0.238 $\pm$ 0.009 & \cellcolor{lightgray}Anthropic Opus (V) & \cellcolor{lightgray}0.285 $\pm$ 0.010  \\
    & \cellcolor{lightgray}Claude: Opus (V) & \cellcolor{lightgray}0.237 $\pm$ 0.012 &  \cellcolor{white}GPT-3.5 (TS-Log-p) & \cellcolor{white}0.295 $\pm$ 0.008   \\
    & \cellcolor{white}Claude: Haiku (Seq) & \cellcolor{white}0.284 $\pm$ 0.012 & \cellcolor{lightgray}Anthropic Haiku (V) & \cellcolor{lightgray}0.297 $\pm$ 0.010   \\
    & \cellcolor{lightgray}GPT-3.5 Turbo (V) & \cellcolor{lightgray}0.292 $\pm$ 0.012 & \cellcolor{white}GPT-4 (TS-Log-p) & \cellcolor{white}0.330 $\pm$ 0.010  \\
    & \cellcolor{white}GPT-3.5 Turbo (TS-Log-p) & \cellcolor{white}0.299 $\pm$ 0.009 & \cellcolor{white}Anthropic Haiku (Seq) & \cellcolor{white}0.355 $\pm$ 0.013   \\
    & \cellcolor{white}GPT-3.5 Turbo (Seq) & \cellcolor{white}0.317 $\pm$ 0.012 & \cellcolor{white}Llama 3 70B (Seq) & \cellcolor{white}0.359 $\pm$ 0.011  \\
    & \cellcolor{white}Llama 3 70B (Seq) & \cellcolor{white}0.316 $\pm$ 0.013 &  \cellcolor{white}Llama 3 70B (TS-Log-p) & \cellcolor{white}0.430 $\pm$ 0.012   \\
    & \cellcolor{white}GPT-3.5 Turbo (Log-p) & \cellcolor{white}0.339 $\pm$ 0.013 & \cellcolor{white}GPT-3.5 (Log-p) & \cellcolor{white}0.473 $\pm$ 0.012  \\
    & \cellcolor{white}Llama 3 70B (TS-Log-p) & \cellcolor{white}0.416 $\pm$ 0.014 &  \cellcolor{white}Anthropic Opus (Seq) & \cellcolor{white}0.483 $\pm$ 0.010  \\
    & \cellcolor{white}Llama 3 70B (Log-p) & \cellcolor{white}0.460 $\pm$ 0.014 & \cellcolor{white}Llama 3 70B (Log-p) & \cellcolor{white}0.525 $\pm$ 0.011  \\
    & \cellcolor{white}GPT-4 (Log-p) & \cellcolor{white}0.507 $\pm$ 0.015 & \cellcolor{white}GPT-4 (Log-p) & \cellcolor{white}0.682 $\pm$ 0.010  \\
    \hline
    
    \multirow{18}{*}{\rotatebox{90}{\textbf{Few Shot Steering}}}
    & \multicolumn{2}{c|}{\cellcolor{white}\textbf{OQA}} & \multicolumn{2}{c|}{\cellcolor{white}\textbf{NYT}} \\
    \cline{2-5}
    & \cellcolor{white}\textbf{Model} & \cellcolor{white}$\mathcal{A}(Y,\hat{Y}_{\mathcal{S}, \mathcal{O}})$ & \cellcolor{white}\textbf{Model} & \cellcolor{white}$\mathcal{A}(Y,\hat{Y}_{\mathcal{S}, \mathcal{O}})$ \\
    \cline{2-5}
    & \cellcolor{lightgray}Anthropic Opus (V) & \cellcolor{lightgray}0.146 $\pm$ 0.007 & \cellcolor{lightgray}Anthropic Opus (V) & \cellcolor{lightgray}0.207 $\pm$ 0.006   \\
& \cellcolor{lightgray}GPT-4 (V) & \cellcolor{lightgray}0.179 $\pm$ 0.008 & \cellcolor{lightgray}GPT-4 (V) & \cellcolor{lightgray}0.208 $\pm$ 0.006  \\
& \cellcolor{lightgray}Anthropic Haiku (V) & \cellcolor{lightgray}0.200 $\pm$ 0.010 & \cellcolor{lightgray}Llama 3 70B (V) & \cellcolor{lightgray}0.208 $\pm$ 0.006   \\
& \cellcolor{white}GPT-4 (TS-Log-p) & \cellcolor{white}0.200 $\pm$ 0.007 & \cellcolor{lightgray}Anthropic Haiku (V) & \cellcolor{lightgray}0.221 $\pm$ 0.006   \\
& \cellcolor{white}GPT-4 (Seq) & \cellcolor{white}0.204 $\pm$ 0.009 & \cellcolor{lightgray}GPT-3.5 (V) & \cellcolor{lightgray}0.226 $\pm$ 0.007  \\
& \cellcolor{lightgray}Llama 3 70B (V) & \cellcolor{lightgray}0.211 $\pm$ 0.011 & \cellcolor{white}GPT-3.5 (Seq) & \cellcolor{white}0.228 $\pm$ 0.007  \\
& \cellcolor{white}Anthropic Opus (Seq) & \cellcolor{white}0.248 $\pm$ 0.010 & \cellcolor{white}GPT-4 (Seq) & \cellcolor{white}0.243 $\pm$ 0.006  \\
& \cellcolor{white}Anthropic Haiku (Seq) & \cellcolor{white}0.262 $\pm$ 0.012 &  \cellcolor{white}Anthropic Haiku (Seq) & \cellcolor{white}0.248 $\pm$ 0.007   \\
& \cellcolor{lightgray}GPT-3.5 (V) & \cellcolor{lightgray}0.280 $\pm$ 0.013 & \cellcolor{white}GPT-3.5 (TS-Log-p) & \cellcolor{white}0.267 $\pm$ 0.007  \\
& \cellcolor{white}GPT-3.5 (TS-Log-p) & \cellcolor{white}0.299 $\pm$ 0.011 &  \cellcolor{white}GPT-4 (TS-Log-p) & \cellcolor{white}0.285 $\pm$ 0.008  \\
& \cellcolor{white}Llama 3 70B (Seq) & \cellcolor{white}0.300 $\pm$ 0.012 & \cellcolor{white}Llama 3 70B (Seq) & \cellcolor{white}0.305 $\pm$ 0.008  \\
& \cellcolor{white}GPT-3.5 (Seq) & \cellcolor{white}0.339 $\pm$ 0.015 & \cellcolor{white}Anthropic Opus (Seq) & \cellcolor{white}0.334 $\pm$ 0.009   \\
& \cellcolor{white}Llama 3 70B (TS-Log-p) & \cellcolor{white}0.394 $\pm$ 0.015 & \cellcolor{white}GPT-3.5 (Log-p) & \cellcolor{white}0.517 $\pm$ 0.010  \\
& \cellcolor{white}Llama 3 70B (Log-p) & \cellcolor{white}0.444 $\pm$ 0.014 & \cellcolor{white}Llama 3 70B (TS-Log-p) & \cellcolor{white}0.593 $\pm$ 0.013  \\
& \cellcolor{white}GPT-4 (Log-p) & \cellcolor{white}0.491 $\pm$ 0.014 & \cellcolor{white}Llama 3 70B (Log-p) & \cellcolor{white}0.629 $\pm$ 0.011  \\
& \cellcolor{white}GPT-3.5 (Log-p) & \cellcolor{white}0.519 $\pm$ 0.016 & \cellcolor{white}GPT-4 (Log-p) & \cellcolor{white}0.650 $\pm$ 0.010  \\
    \hline
    \multirow{18}{*}{\rotatebox{90}{\textbf{ }}} 
    & \multicolumn{2}{c|}{\cellcolor{white}\textbf{OQA}} & \multicolumn{2}{c|}{\cellcolor{white}\textbf{NYT}} \\
    \cline{2-5}
    & \cellcolor{white}\textbf{Model} & \cellcolor{white}$\mathcal{A}(Y,\hat{Y}_{\mathcal{S}, \mathcal{O}})$ & \cellcolor{white}\textbf{Model} & \cellcolor{white}$\mathcal{A}(Y,\hat{Y}_{\mathcal{S}, \mathcal{O}})$ \\
    \cline{2-5}
    & \cellcolor{white}Discretization Error (Seq) & \cellcolor{white}0.136 $\pm$ 0.012 &  \cellcolor{white}Discretization Error (Seq) & \cellcolor{white}0.115 $\pm$ 0.001 \\
    
    & \cellcolor{white}Uniform & \cellcolor{white}0.381 $\pm$ 0.009 & \cellcolor{white}Uniform & \cellcolor{white}0.223 $\pm$ 0.006 \\
    
    & \cellcolor{white}Majority Vote & \cellcolor{white}0.731 $\pm$ 0.017 & \cellcolor{white}Majority Vote & \cellcolor{white}0.700 $\pm$ 0.008 \\

 & \cellcolor{white}Minority Vote & \cellcolor{white}0.837 $\pm$ 0.014 & \cellcolor{white}Minority Vote & \cellcolor{white}0.845 $\pm$ 0.007 \\

    \hline
    \caption{\textbf{Comparing model performance across OQA and NYT datasets and steering methods.} Models are ranked based on mean total variation and highlighted in gray and with (V) have a distribution expression method of directly verbalizing the distribution in a JSON format ($\mathcal{O} = \texttt{Verbalize}$). Models not highlighted represent samplers, where (Seq) represents the 30-token sequential distribution output ($\mathcal{O} = \texttt{Sequence}$), (Log-p) represents $\mathcal{O} = \texttt{Model Log-probabilities}$, and (TS-Log-p) represents $\mathcal{O} = \texttt{Temperature Scaled Model Log-probabilities}$.}
    \label{tab:all_results}
\end{longtable}

\twocolumn